%% file: example_paper.tex
\documentclass{article}

\usepackage{microtype}
\usepackage{graphicx}
\usepackage{booktabs} 

\usepackage{hyperref}
\usepackage{caption}
\usepackage{subcaption}
\usepackage{booktabs}
\usepackage{colortbl}
\usepackage{wrapfig}

\usepackage[accepted]{icml2024}

\usepackage{amsmath}
\usepackage{amssymb}
\usepackage{mathtools}
\usepackage{amsthm}
\usepackage{adjustbox}

\usepackage[capitalize,noabbrev]{cleveref}

\theoremstyle{plain}

\theoremstyle{definition}

\theoremstyle{remark}

\usepackage[textsize=tiny]{todonotes}

\icmltitlerunning{}

\begin{document}

\twocolumn[

\icmltitle{Overconfident Oracles: Limitations of In Silico Sequence Design Benchmarking}





\icmlsetsymbol{equal}{*}

\begin{icmlauthorlist}
\icmlauthor{Shikha Surana}{InstaDeep}
\icmlauthor{Nathan Grinsztajn}{InstaDeep}
\icmlauthor{Timothy Atkinson}{InstaDeep}
\icmlauthor{Paul Duckworth}{InstaDeep}
\icmlauthor{Thomas D.\ Barrett}{InstaDeep}
\end{icmlauthorlist}

\icmlaffiliation{InstaDeep}{InstaDeep, London UK}

\icmlcorrespondingauthor{Shikha Surana}{s.surana@instadeep.com}

\icmlkeywords{Machine Learning, ICML}

\vskip 0.3in
]



\printAffiliationsAndNotice{}  

\begin{abstract}
Machine learning methods can automate the \textit{in silico} design of biological sequences, aiming to reduce costs and accelerate medical research. Given the limited access to wet labs, \textit{in silico} design methods commonly use an oracle model to evaluate \textit{de novo} generated sequences.
However, the use of different oracle models across methods makes it challenging to compare them reliably, motivating the question: are \textit{in silico} sequence design benchmarks reliable?
In this work, we examine 12 sequence design methods that utilise ML oracles common in the literature and find that there are significant challenges with their cross-consistency and reproducibility. Indeed, oracles differing by architecture, or even just training seed, are shown to yield conflicting relative performance with our analysis suggesting poor out-of-distribution generalisation as a key issue.
To address these challenges, we propose supplementing the evaluation with a suite of biophysical measures to assess the viability of generated sequences and limit out-of-distribution sequences the oracle is required to score, thereby improving the robustness of the design procedure.
Our work aims to highlight potential pitfalls in the current evaluation process and contribute to the development of robust benchmarks, ultimately driving the improvement of \textit{in silico} design methods. 
\end{abstract}

\section{Introduction}\label{sec:introduction}

Utilising generative machine learning (ML) to design biological sequences that maximise desired properties, such as binding affinity or expression level, is important for advancing the medical research field and its applications. Experiments conducted \textit{in vitro} are expensive and time-consuming, and thus, leveraging ML to automate the \textit{in silico} design of sequences to have a high likelihood of \textit{in vitro} success can reduce costs and accelerate research progress.

In recent years, numerous \textit{in silico} design methods have been proposed to generate biological sequences \cite{coms, gflownets, pex, chen2023bidirectional, bootgen}. Typically, biological sequences can be 100's to 1000's of characters long, leading to vast search spaces, but often have access only to a limited dataset of example sequences and their ground-truth values. As a result, to evaluate sequence design methods, an ML oracle model is often trained on the dataset and used to score \textit{de novo} generated sequences, thereby simulating the wet lab evaluation. However, within the community there is a general lack of consensus on the specific oracle model parameters and architecture, which results in the use of different oracle models across different studies. 
It is common practice to propose novel sequence design methods alongside a tailored evaluation pipeline, including the choice of oracle. 
Inconsistency across oracle models hinders any reliable comparisons of design methods, and brings into question the robustness of \textit{in silico} sequence design benchmarks. 

\textbf{Contributions} Our first contribution is to investigate the reliability of 12 design methods. Specifically, whether superfluous changes to the oracle can impact the relative performance of methods and affect their overall ranking when scored against: (1) the same oracle architecture trained with five different seeds and (2) three different oracle architectures. 
We perform this on two different tasks: 5' untranslated region (UTR, DNA task) \cite{utr_pjsample}, and green fluorescence protein (GFP, protein task) \cite{gfp_sarkisyan}. 
We demonstrate high variance and a lack of consensus in the methods' relative performance and present insights suggesting that issues arise from the oracle's poor out-of-distribution (OOD) generalisation.

Our second contribution introduces a suite of biophysical measures, specifically tailored for DNA and protein sequence design tasks, to assess the biological validity of \textit{de novo} generated sequences. We demonstrate that these measures are critical due to ML oracle's \romannumeral1) poor OOD generalization, which necessitates reducing OOD sequences being evaluated and improving reliability, and \romannumeral2) lack of biological knowledge, which prevents it from filtering out biologically unfit sequences.

Our work complements the growing interest within the community towards improving \textit{in silico} benchmarks for biological tasks. Recent studies have proposed biophysical measures to improve the benchmarks for \textit{de novo} structure-based \cite{posebusters, posecheck} and sequence-based \cite{frey2023protein, spinner2024well} design tasks. 
In our work, we highlight critical limitations of the current \textit{in silico} protein and DNA sequence design benchmarks, and further, introduce additional biophysical measures to improve the robustness and reliability. 

\section{Related Work}\label{sec:related_work}

\textbf{Sequence Design Methods and Tasks}
Several works have developed offline methods to tackle the problem of sequence design, particularly through reinforcement learning \cite{angermueller2019model}, population-based optimisation \cite{angermueller2020population}, model-based optimisation \cite{coms, chen2023bidirectional}, deep generative models \cite{mins_kumar, gflownets, bootgen}, and evolutionary search \cite{pex}.
To help provide a level of standardisation for biological sequence design tasks and methods, there has recently been several open-source resources, for tasks:  ProteinGym \cite{notin2024proteingym}, DesignBench \cite{designbench_trabucco}, and FLEXS\footnote{\url{https://github.com/samsinai/FLEXS}}, and for methods: Design Baselines \cite{coms}. Our experiments include some of the methods mentioned above, as well as Design Bench and Design Baselines suites.

\textbf{Evaluation of \textit{in silico} Benchmarks} A recent, important area of research is assessing the physical and chemical plausibility of ML-generated solutions for biological tasks. Previous studies demonstrate that ML-based methods tend to generate physically implausible structures for tasks such as docking \cite{posebusters} and structure-based drug design \cite{posecheck}, and these works present a suite of biophysical measures to validate the biological viability of generated complexes. Similarly, prior works have presented biologically-inspired measures for protein design tasks \cite{frey2023protein, spinner2024well}. Our work expands the scope to include both protein and DNA tasks, and introduces additional measures for each task setting that are evaluated against both task and external datasets. 

\textbf{Generalisation of Sequence-Scoring Models} The work by \citet{tagasovska2023antibody} discusses how surrogate models can fail to identify true casual and mechanistic links between (parts of) the sequences and the biological property of interest which is scored, leading to poor generalisation to unseen sequences. The findings of this study raise questions about the oracle's generalization ability and we investigate this in our work.

\section{Experimental Setup}
In this section we outline the biological sequence design datasets, tasks (including the oracle models), and the generators that will be used in the subsequent experiment sections. 

\subsection{Datasets} 
This work considers three datasets: green fluorescence protein (GFP), 5' untranslated region (UTR), and transcription factor binding sequences of length 8 (TFBind-8), each consisting of sequences annotated with a particular biological characteristic of interest. The distribution of ground-truth scores for each task is presented in Appendix Figure~\ref{fig:score_distributions}.

\textbf{GFP} dataset contains protein sequences of length 237, consisting of 20 possible amino acids, annotated with a ground-truth value corresponding to its fluorescence level. It is curated by \citet{gfp_sarkisyan} and comprises $51,715$ unique sequences. Each sequence in this dataset has up to $15$ mutational edits, with an average of $3.7$ edits compared to the wild-type sequence. 

\textbf{UTR} dataset contains DNA sequences of length 50, constructed using 4 nucleobases: adenine (A), guanine (G), cytosine (C), and thymine (T). It consists of $280,000$ sequences~\citep{utr_pjsample}, each annotated with their ribosome loading, which is correlated with the expression level of the 5'UTR region.

\textbf{TFBind-8} dataset contains sequences of length 8, consisting of four distinct nucleobases, each annotated with a ground-truth value of binding activity with human transcription factors. The dataset consists of $65,536$ sequences, i.e. all possible sequences of four nucleobases of length 8, i.e $4^8$.

\subsection{Tasks}\label{subsec:oracles}
The biological datasets described above can be formulated into sequence design tasks where the aim is to design sequences that maximise a desired property (which is typically expressed as the score of the sequences). Additionally, these tasks include an oracle model responsible for scoring \textit{de novo} sequences generated by a design method.

For UTR and GFP datasets, the oracle employed is usually an ML model trained on the available dataset of sequences and their corresponding fitness values. In the following we describe the commonly used oracles for these tasks. In contrast, TF-Bind dataset provides values for every possible sequence, so there is no need for an oracle model, as \textit{de novo} sequences are queried against the dataset. 

\textbf{GFP} Several oracle models are commonly used for GFP, we consider the following three: 
1) Design Bench Transformer \cite{designbench_trabucco} used in \citet{coms, gflownets, bootgen}, 
2) TAPE \cite{tape} used in \citet{pex, song2023importance, wang2023self}, and 3) ESM-1b \cite{esm1b_rives} with a fine-tuned head trained and used in \citet{pex}. We train the Trasnforer oracle ourselves (according to the Design Bench implementation), and use the available pre-trained checkpoints for TAPE and ESM-1b models.
The Transformer oracle is trained on a random uniform $90/10\%$ train/validation split of the task dataset. TAPE and the ESM-1b are trained on task dataset sequences that are $3$ mutational edits away from the wild-type sequence, and the remaining sequences of $4-15$ mutations constitute the validation set.  

\textbf{UTR} There are two popular oracle architectures used for UTR: convolutional neural network (CNN) \cite{utr_pjsample, angermueller2020population}, and residual neural network (ResNet) \cite{coms, bootgen}. Following recent works, we use the ResNet oracle and employ the Design Bench parameters and training code~\cite{designbench_trabucco}. The ResNet oracle is trained on approximately $93\%$ of the dataset ($260,000$ sequences) and is validated on the remaining $7\%$. 

\textbf{TFBind-8} This task has a ground-truth oracle as it includes a fully enumerated dataset that can be queried to retrieve the experimental scores of each sequence.

\subsection{Sequence Generators}\label{subsec:methods} 
In this work, we evaluate 12 well-known design methods developed for biological sequence design: 
1) Generative Flow Networks with Active Learning (GFN-AL\footnote{\url{https://github.com/MJ10/BioSeq-GFN-AL}}, \citet{gflownets}),
2) Bootstrapped training of score conditioned Generator (BootGen\footnote{\url{https://github.com/kaist-silab/bootgen}}, \citet{bootgen})
3) Conditioning by Adaptive Sampling (CbAS, \citet{cbas_brookes}), 
4) Autofocused CbAS (Auto. CbAS, \cite{autofocused_fannjiang}), 
5) Bayesian optimization with a quasi-expected improvement acquisition function (BO-qEI, \citet{wilson2018maximizing}), 
6) Model Inversion Networks (MINs, \citet{mins_kumar}), 
7) Covariance Matrix Adaptation Evolution Strategy (CMA-ES, \citet{hansen2006cma}), 
8) REINFORCE \cite{reinforce}, 
9)-11)  Gradient Ascent (GA) with respect to a surrogate model, including two variations - taking the mean (GA Mean) and the minimum (GA Min) of the ensemble \cite{designbench_trabucco}, 
12) and Conservative Objective Models (COMs, \citet{coms}).
Implementations from Design Baselines repository\footnote{\url{https://github.com/brandontrabucco/design-baselines}}. 

After training, all methods are sampled to obtain a batch of 128 sequences. When performing seeded runs, each method is retrained and then sampled.

\input{tabs/experiment1}

\section{Evaluating \textit{de novo} sequences with ML oracles}

Our first contribution is to highlight the limitations in the evaluation process when leveraging ML-trained oracles for biological \textit{de novo} sequence design. 

As a practitioner, it is of utmost importance to have confidence in the oracles employed to provide ground truth values such that, ultimately, only the most promising \textit{in silico} \textit{de novo} sequences are proposed for (potentially expensive and time-consuming) evaluation \textit{in vitro}. 
One question we ask here is whether the relative performance of different design methods vary, as we vary superfluous characteristics of the oracle. 
For example, are the leading design methods consistently performant with oracles trained across many random seeds or different architectures?

In Section \ref{subsec:rankings}, we reveal that the relative performance of 12 commonly used sequence design methods is highly sensitive to both (1) the random seed used to train the oracle and (2) the architecture of the oracle employed to score new sequences. 
These findings cast doubts on the conclusions of prior works, as it becomes challenging to determine whether a method is genuinely state-of-the-art or simply outperforms other methods due to inherent randomness in a specific oracle implementation.
In Section \ref{subsec:generalisation}, we dive into the reasons behind these inconsistencies, offering insights that suggest the poor generalization capabilities of commonly used ML oracle models may be a contributing factor.

\subsection{What is state-of-the-art?}\label{subsec:rankings}

Many prior works that propose sequence design methods do so by training their own ML oracle to evaluate new sequences generated. 
They often leverage open-source implementations or implement their own architecturally different oracle which is trained on open-source datasets.
As an example of the former, Design Bench open-source code for training oracles but not the weights of the oracle model itself, resulting in each study re-training their own oracle model. 
Since the Design Bench implementation is not seeded, each study ends up with different oracle model parameters. 
As an example of the latter, prior works use one of four architecturally different oracles (Design Bench Transformer \cite{coms, bootgen}, TAPE \cite{pex}, and ESM-1b \cite{pex}) for the GFP task in prior works. Potential inconsistencies amongst the oracles may cause unreliable comparisons between methods. 

Since many different oracles are used in prior works for any given sequence design task, minor variations in evaluated scores are to be expected. However, what we would expect, is low-variance across the highest scored sequences (across batches), and that under each oracle, there is a consistent ranking of relative performance of each design method. 
To examine this, we test the consistency of the design methods evaluated against: 
(Experiment~1) a single oracle trained across five different random seeds, and 
(Experiment~2) three different ML oracle architectures (trained on the same task).

\textbf{Experiment 1} To assess how robust the evaluation of design methods is under ML-based oracles, we select one oracle setup and vary the random seed used during training. If the oracle models were reliable and consistent, we would not expect this to affect the relative performance of the design methods. 
We demonstrate this on two sequence design tasks, UTR and GFP, and utilize the Design Bench ML oracles. Specifically, we re-train the oracles with five different random seeds, to assess how these random replications affect the performance of each design method. 
We train each of our 12 design methods on 8 random seeds as described in Section \ref{subsec:methods}, and sample a batch of 128 sequences from each. 

\textbf{Results 1} In Table \ref{tab:gfp_seeds} we present the ranking of the maximum score achieved from sampled \textit{de novo} batches for the 12 sequence design methods (averaged over 8 random seeds), where each column is a different seeded replication of the Design Bench oracle for (a) UTR and (b) GFP tasks. 
Table~\ref{tab:methods_robustness} (Appendix) shows the rankings of the 12 design methods for each of the 8 random seeds (columns) under a single oracle model instance. 
We collate the results in Figure~\ref{fig:method_score_distribution} (Appendix) which shows the distribution of each design method's maximum sequence score under 5 seeded oracle models and 8 seeded designs. 

For the GFP task, we see that three different design methods are considered SOTA in Table \ref{tab:gfp_seeds}(b).
Additionally, when considering the ranking of the 8 seeded design methods scored under a single oracle instance, we see from Table \ref{tab:methods_robustness} that the results are highly variable, and there is a general lack of agreement on the relative performance across design seeds.
Notably, for UTR, GFN-AL generated the highest scoring sequences under two seeds and the worst under the remaining 6 seeds. 
When aggregated across the 8 seeds in Table \ref{tab:gfp_seeds}, we see some agreement among the different replications of the oracle models. Specifically, the methods that consistently generate low-scoring sequences are ranked in the bottom three, and for the UTR task, BootGen consistently generates SOTA \textit{de novo} sequences. However, for the latter, we demonstrate in the following section that this is because BootGen is able to exploit a key limitation of ML-based oracles which is that they are unreliable in scoring sequences outside of their train set distribution. 
\textbf{This highlights our first contribution: evaluating design methods based on approximate, self-managed oracles, does not lead to insights into the design methods themselves, but rather the randomness inherent in the oracle evaluations. }

\textbf{Experiment 2} 
To assess whether there is consistency among the relative performance of the 12 design methods when utilizing ML oracles, we compare oracles with different architectures (each optimised for the same task). 
We demonstrate this on the GFP sequence design task using 3 oracle models: (1) Design Bench Transformer, (2) TAPE, and (3) ESM-1b. 
Again, we generate a new batch of 128 sequences from each design method (under 8 random seeds) and evaluate their scores under each of the oracle models. We then rank the methods based on the maximum oracle score achieved within each generated batch, which is a common metric in the literature. 

\textbf{Results 2} Table \ref{tab:gfp_oracles} presents the rankings of the 12 methods for the three different oracle architectures. Although there is a general consensus that BootGen performs comparatively well and CMA-ES poorly, an overall lack of agreement among the rankings assigned by the different oracle models is evident. 
Noticeably, under ESM-1b GFN-AL and BO-qEL generate the highest scoring \textit{de novo} sequences, however, under the Design Bench and TAPE oracles, rank these sequences as the tenth best, and BO-qEI as the lowest-scoring method. 
Similarly, MINs is the third best method under Design Bench, fifth best under TAPE, and eighth best under ESM-1b. 
\textbf{Our second takeaway is that allowing the community to evaluate generative sequence design methods by utilising different approximate ML oracles can be highly subjective to specific architectural choices. }

\input{tabs/gfp_experiment2}

\textbf{Concluding Remarks} The inconsistency between design methods' relative performance across superfluous oracle-design choices highlights the potential pitfalls of relying on ML-based oracle for rigorous evaluation. Our results demonstrate that the choice of the oracle model heavily influences the perceived state-of-the-art performance across both DNA and protein sequence design tasks, emphasising the importance of either considering multiple oracles and/or including additional evaluation metrics for a more comprehensive and reliable assessment of \textit{de novo} sequences. 

\subsection{Do ML oracles generalise?}\label{subsec:generalisation}

We hypothesise that a major contributing factor to the variation of design methods' performance across different ML oracles, suggests inadequacies in the oracles themselves. Specifically, in their ability to generalize out-of-distribution (OOD) i.e. to new \textit{de novo} sequences outside of the training data. 
In this section, we first evaluate the generalisation capabilities of three commonly used GFP sequence design oracles. 
Our results reveal poor generalisation performance across all oracle models. 
Consequently, we delve into the oracle training procedure to better understand the underlying reasons for the observed limitations.

\textbf{Experiment 3} We analyse the performance of three commonly used ML oracles trained to score and evaluate \textit{de novo} GFP protein sequence designs. We evaluate for both in-distribution and out-of-distribution error by utilizing the ground truth scores from the corresponding datasets. The three oracles are: 1) Design Bench Transformer, 2) TAPE, and 3) ESM-1b. 
Specifically, we investigate the error between the oracle's predicted score and the true experimental score taken directly from the dataset for both train and held-out validation data splits. 
Clearly, one expects low training set error, and understandably higher validation set error.

\begin{figure*}[t!]
    \centering
    \includegraphics[width=0.32\linewidth]{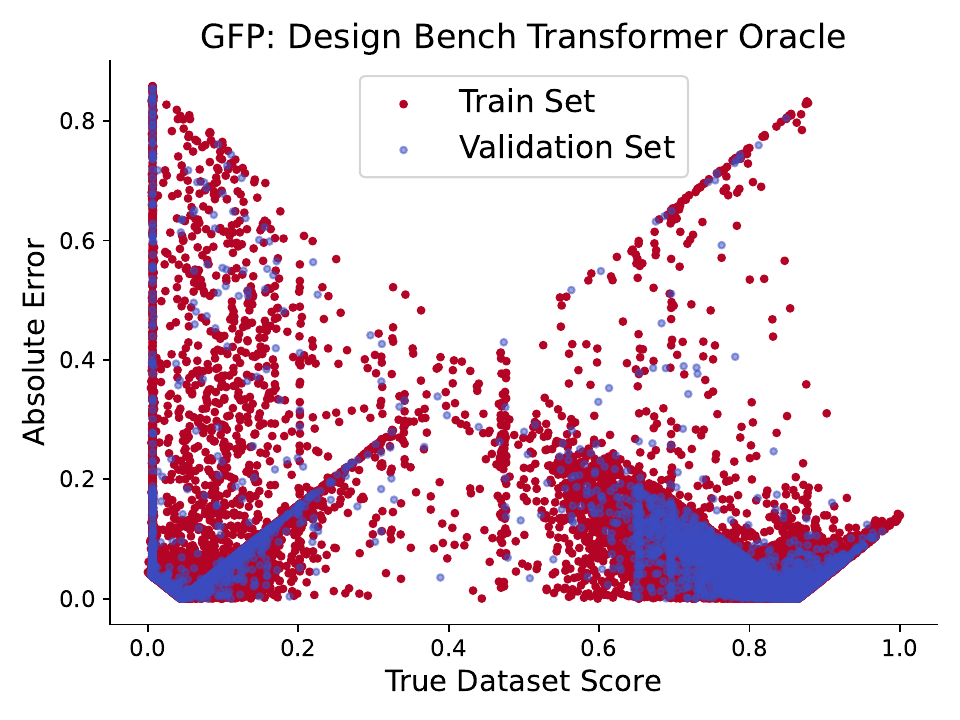}
        \includegraphics[width=0.32\linewidth]{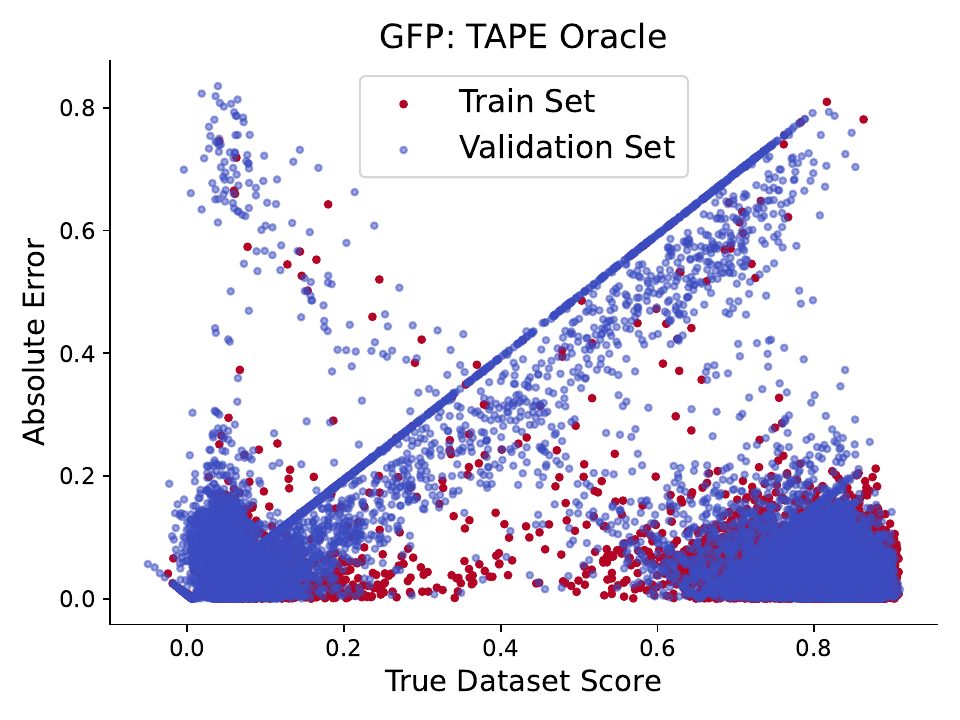}
    \includegraphics[width=0.32\linewidth]{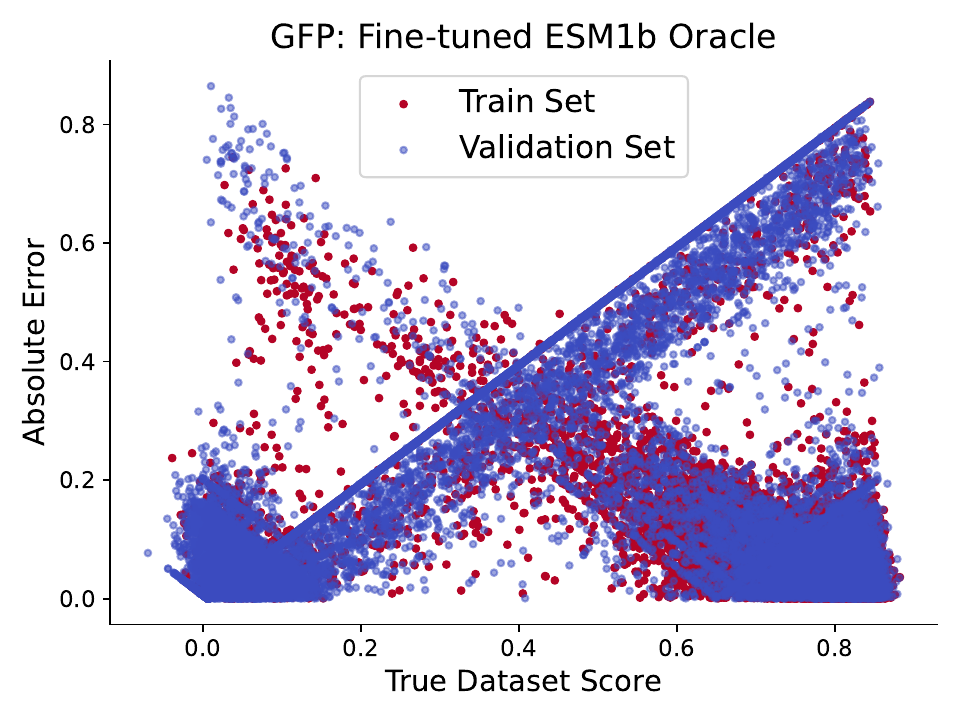}
    \caption{
    Error between the oracle predictions and true dataset scores per sequence for the train (red) and held-out validation (blue) datasets for three GFP oracles: 
    (left) Design-Bench transformer,
    (middle) TAPE, 
    (right) ESM-1b fine-tuned on GFP dataset.}
    \label{fig:mse_oracle_true_score_gfp}
\end{figure*}

\textbf{Results 3} Figure \ref{fig:mse_oracle_true_score_gfp} presents the absolute error between each oracle scored sequence and the true dataset score for Design Bench Transformer, TAPE, and ESM-1b oracles. 
The Design Bench oracle demonstrates poor accuracy for both train and held-out validation sequences, suggesting that the oracle struggles to fit the training dataset accurately. 
Despite exhibiting low errors for some held-out sequences, the oracle overall does not generalise well to both \textit{de novo} sequences, and sequences it has seen before. 

The TAPE oracle achieves reasonable accuracy on the training dataset sequences; however, it shows poor generalisation to the held-out validation sequences. Considering the train/validation split (described in Section~\ref{subsec:oracles}), TAPE fits the training dataset (sequences of up to 3 mutations from the GFP wildtype) more accurately, however, we see that it struggles to effectively generalize to the validation sequences (with greater than 3 mutations). 

Finally, ESM-1b exhibits poor accuracy for sequences in both the train and validation sets, indicating a similar issue as observed with the Design Bench model: the model fails to fit the training dataset, and should not be relied on to provide robust ground truth scores for held-out sequences. 

\textbf{Analysis} Overall, we find that all 3 commonly used oracles demonstrate poor generalisation capabilities on the GFP design task. 
It is interesting to note the similarity in the error distribution shown across all oracles. To better understand this, we highlight the data distribution of the available GFP ground truth scores in Appendix Figure~\ref{fig:score_distributions} (centre). 

The distribution is extremely unbalanced and bi-modal: one mode between 0 and 0.2 encompassing approximately 40\% of the data; and the second mode between 0.6 to 0.95 representing roughly 60\% of the data. 
In light of this unbalanced data distribution, and the common practice of taking random data splits, one hypothesis is that the ML oracles converge to simply predicting the score of every sequence to one of the two modes. Consequently, the error plots reveal two distinct peaks, each reflecting increasing error as the sequences vary from these two modes. 

\subsection{Analysing generalisation via state space coverage}

Biological sequence design tasks typically have a combinatorially large search space that grows exponentially with the sequence length.  By contrast, the datasets available for many of these tasks cover only a tiny fraction of this space.

We investigate this phenomenon using the UTR task, with a sequence length of 50, and therefore state-space of $4^{50}$ possible combinations. The available UTR labelled dataset is 280,000 sequences representing less than $0.001\%$ of the total possible space. 
In the previous section, we demonstrated the oracle's lack of ability to generalise OOD. 
A resulting outcome of this, is that for example, the Design Bench ML oracle can predict surprisingly high scores (outside of the dataset range) of 0.78 and 0.86 to sequences composed entirely of either adenine (A) or thymine (T) bases. The highest score in the available dataset is 0.73. 

Due to the unavailability of ground truth scores for all sequences in the UTR state space, we cannot fully assess the oracle generalisation to out-of-distribution sequences without further restricting the training dataset. However, we can recreate an equivalent ML oracle on a smaller DNA sequence design task, for example, leveraging the TFBind-8 DNA dataset that spans the entire state space $4^8$ (= 65\,536) sequences.

\textbf{Experiment 4} To faithfully recreate the UTR Design Bench ML oracle using the TFBind-8 dataset, we train an equivalent ResNet oracle on 1\% of the TFBind-8 dataset, with an equivalent random data split strategy. 
(Note 1\% train split is an overestimation as compared to the UTR task data splits). 

\begin{figure}[h!]
    \centering
    \includegraphics[width=0.66\linewidth]{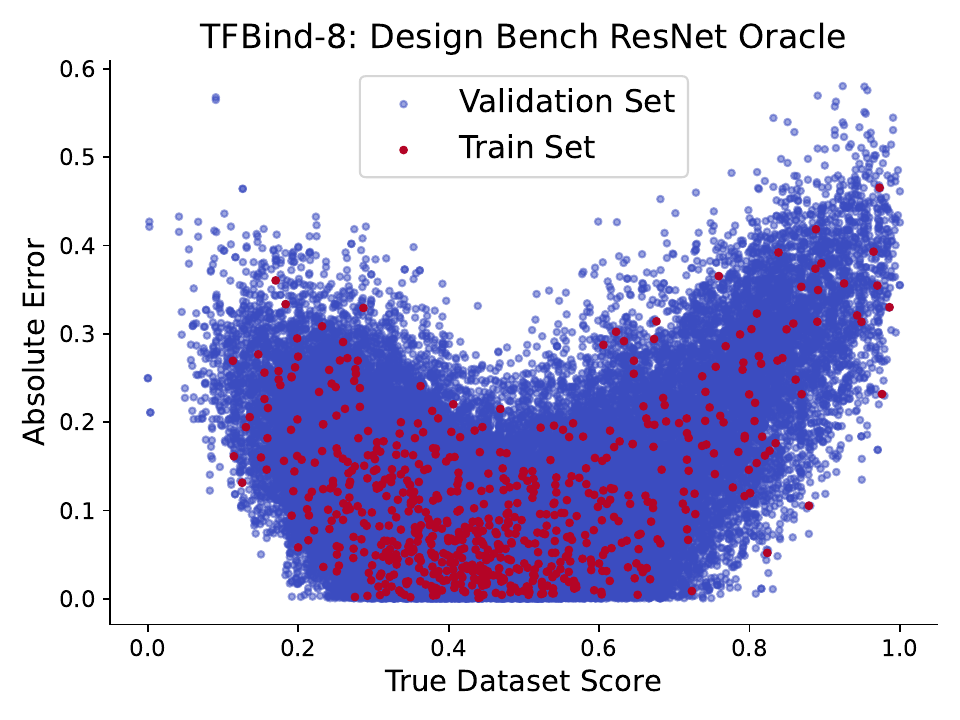}
    \caption{TFBind-8: Absolute error between the Design Bench-inspired ML oracle predictions and ground truth dataset scores for train (red) and held-out validation (blue) datasplits.}
    \label{fig:mse_oracle_true_score}
    \vspace{-2.5ex}
\end{figure}

\textbf{Results 4} Figure \ref{fig:mse_oracle_true_score} illustrates the absolute error between the scores in the dataset and the oracle predicted scores, for both the training dataset (1\% state space coverage) and held-out validation set (remaining 99\% state space coverage). 
Given that TFBind-8 provides experimentally computed ground truth scores directly from a wet lab experiment, (albeit subject to inherent noise), comparing these scores with those predicted by the ML oracle offers insight into how reliably the Design Bench-inspired ML oracle can possibly represent the wet lab from such small state-space coverage. 
Our results highlight a larger error for held-out validation sequences, indicating poor generalisation of the oracle to unseen data. Notably, the oracle exhibits significant errors at the extremes of the score distribution, corresponding to sequences that are either truly high- or low-performing -- which from a design task perspective are the sequences a practitioner would be most interested in robustly scoring \textit{in silico}. 

\textbf{Concluding Remarks} We have demonstrated that within common DNA and protein sequence design tasks, training an ML-based oracle on a tiny fraction, e.g. less than 0.001\%, of the possible space of sequences, has a tendency to cause poor accuracy in scoring sequences beyond its training distribution (and often within!). 
This issue is particularly highlighted with the TAPE oracle for GFP, which fails to generalize even to sequences with 3-15 mutations, let alone the remaining ~230 mutations required to evaluate on entirely new \textit{de novo} generated protein sequences. 
Previous studies often make the assumption that the ML-based oracles accurately represent wet lab scores, and thus, use it to evaluate the sequences generated by their method. However, since some methods do not constrain their generation process to align with the dataset distribution, oracle are forced to score wildly OOD sequences, leading to unreliable evaluations.

\section{Leveraging Biophysical Measures for Improved Sequence Design}

In the previous section, we demonstrated that popular ML oracles struggle to generalise OOD and can assign high scores to seemingly implausible sequences. In this section, we propose a strategy based on the hypothesis that a single ML oracle is not reliable enough to be used in isolation when evaluating a design method due to its aforementioned limitations. 
That is, we introduce a suite of biophysical measures that can assist practitioners in reducing the sequence state space and alleviate the oracle being evaluated OOD on \textit{de novo} generated sequences. 

When practitioners generate a batch of sequences to be sent for evaluation in a wet lab, it is crucial to ensure those sequences are good candidates and biologically sound, to avoid wasting resources. To achieve this, we propose that in practical settings, a suite of biophysical measures can be employed that leverage the data-distributions in the available data. 
By applying these measures to a batch of new sequences, we can automatically identify those with a higher likelihood of being biologically valid and more likely to succeed in wet lab experiments. These measures essentially reduce the search space of possible sequences by grounding the generation to regions associated with available data. 

\subsection{Suite of Biophysical Measures}
The suite of measures represents a set of checks to validate whether sequences are biologically plausible (with a high degree of confidence). Whilst this is not an exhaustive set of biophysical measures one could use, we aimed to include somewhat general measures indicative of biological fitness, for both DNA and protein sequences. 

\textbf{DNA Measures} Three DNA measures are \textit{coverage}, \textit{guanine-cytosine (GC) content}, and \textit{homopolymer}. 
\textit{Coverage} refers to the proportion of each of the four nucleobases represented in the sequence. 
\textit{GC content} measures the percentage of guanine and cytosine bases in a sequence. 
GC content of a sequence can significantly impact thermostability which is a vital aspect of success in a wet lab. 
A \textit{homopolymer} is a stretch of consecutive identical nucleobases in a sequence. Long homopolymers can cause issues in downstream applications, such as PCR and sequencing, and thus, are good indicators of artificial sequences. 

\textbf{Protein Measures} 5 protein measures are \textit{molecular weight}, \textit{aromaticity}, \textit{isoelectric point}, \textit{grand average of hydropathy (gravy)}, and \textit{instability index}. 
\textit{Molecular weight} is the sum of the atomic weights of all atoms in a protein molecule and can influence the protein's stability, folding, and function. 
\textit{Aromaticity} refers to the presence and distribution of aromatic amino acids (tryptophan, tyrosine, and phenylalanine) in a sequence. Aromatic residues play critical roles in protein stability, folding, and interactions with other molecules.
The \textit{isoelectric point} is the pH at which a protein has a net charge of zero (i.e., an equal number of positively and negatively charged residues) and this is a crucial factor in protein solubility. 
\textit{Gravy} is a measure of the overall hydrophobicity or hydropathy of a protein sequence and provides insights into the protein's structural and functional properties.
The \textit{instability index} is a measure of a protein's susceptibility to degradation or denaturation. A high (low) instability index indicates that the protein is likely to be unstable (stable) and have a shorter (longer) half-life. 

\subsection{Evaluating with Biophysical Measures}

\textbf{Evaluation Procedure} For each biological measure, we introduce an acceptable range under a reference dataset as the 99$\%$ middle quantile, and consider a sequence valid if it lies within that range for all measures. 
During our work, we considered two  alternatives for the reference dataset: (1) the specific task dataset, and 
(2) a general distribution of natural sequences. 
Both strategies provide a data distribution from valid and biologically plausible sequences taken from the available data that we can leverage to reduce \textit{de novo} generation of implausible sequences. 

To incorporate these measures into the evaluation of our sequence design methods, we sample a batch of 128 sequences and evaluate each sequence with respect to the suite of measures to determine whether it is likely to be a plausible sequence. We report the percentage of valid sequences for each design method, where clearly, higher is better.

\begin{figure*}[t!]
    \centering
    \includegraphics[width=0.33\linewidth]{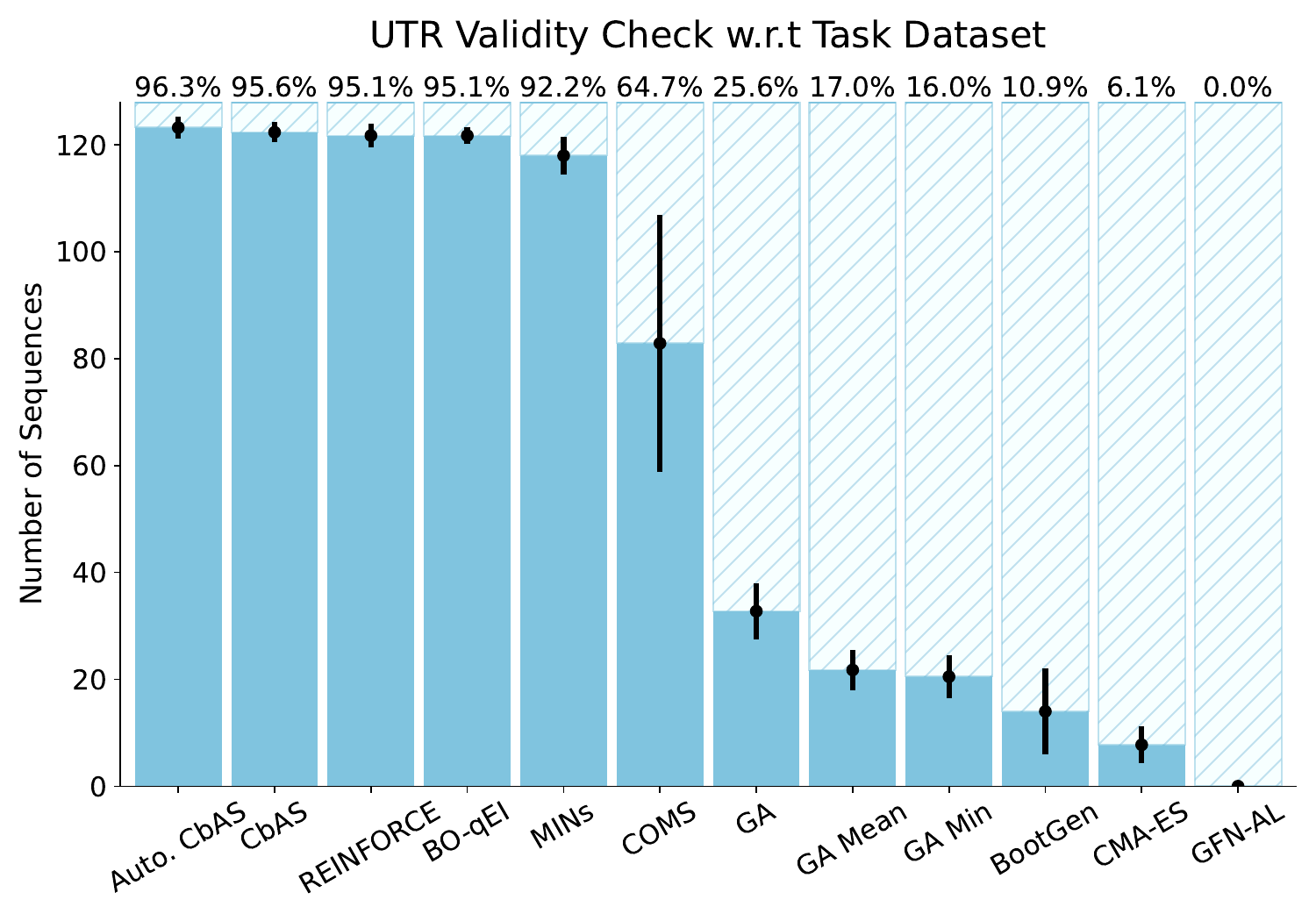}
    \includegraphics[width=0.33\linewidth]{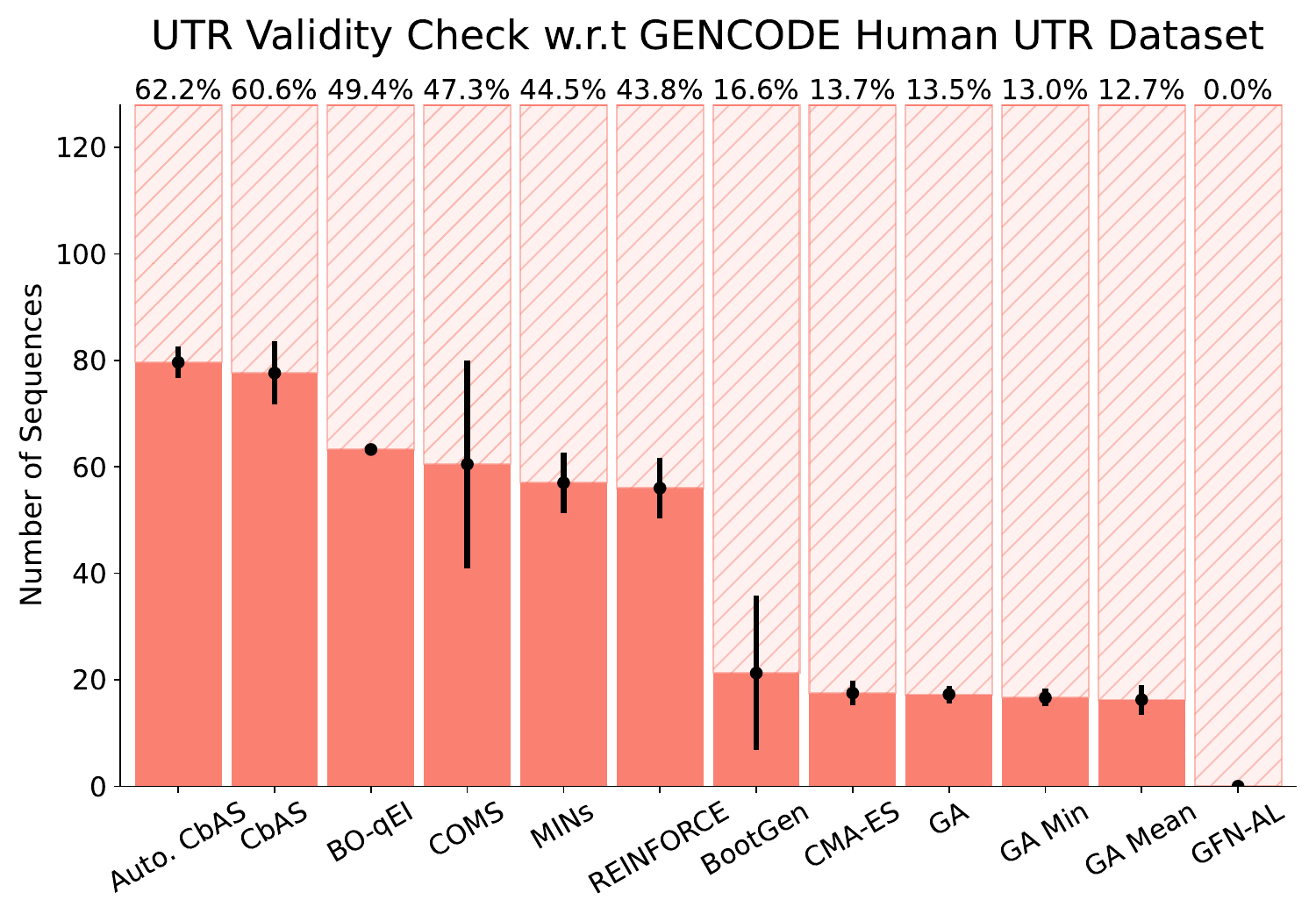}
    \includegraphics[width=0.33\linewidth]{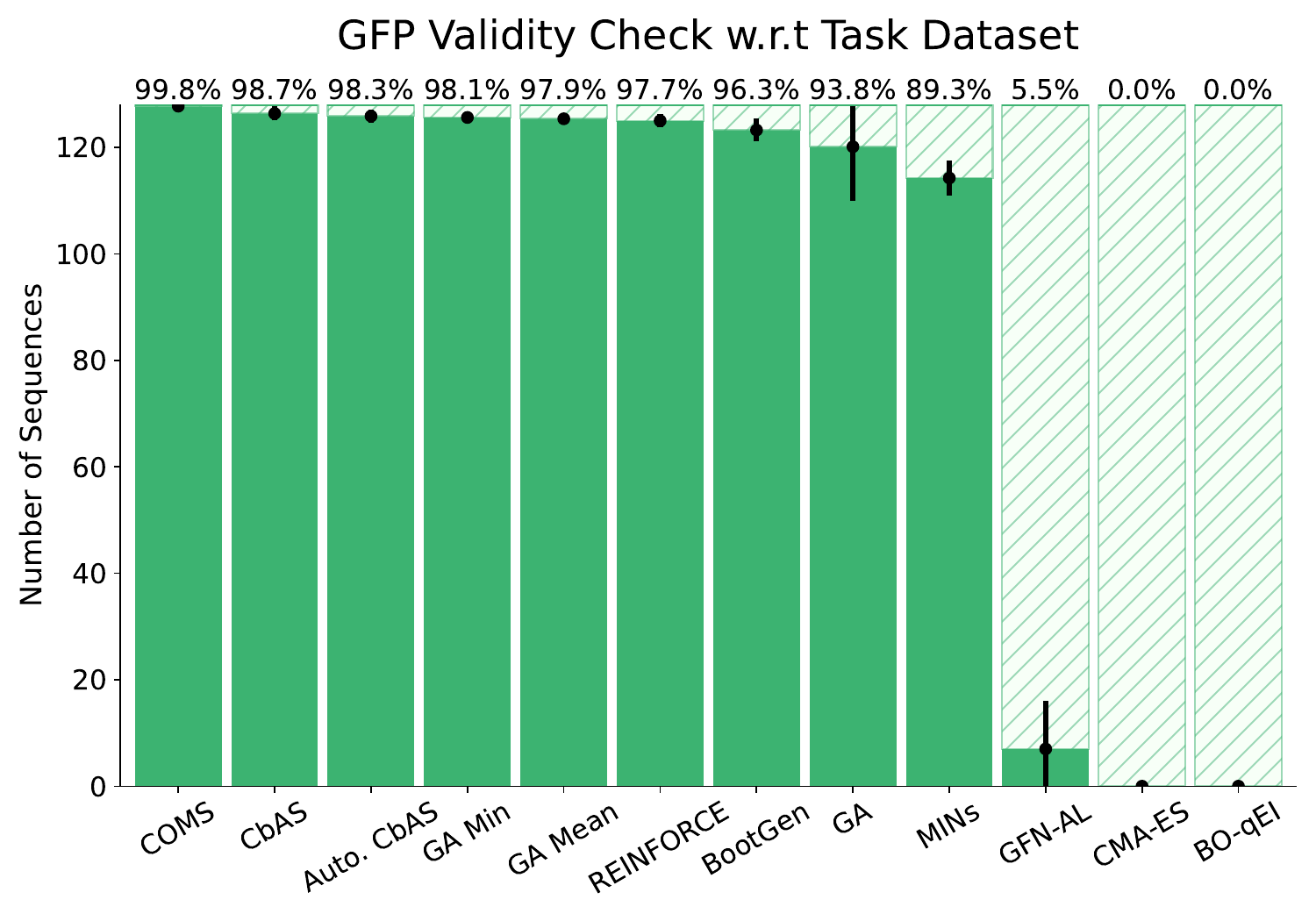}
    \caption{
    Generated sequences classified by DNA (left and centre) and protein (right) suite of biological measures, for 12 sequence design methods with respect to the task dataset (left and right) and the GENCODE database (centre).}
    \label{fig:validity}
\end{figure*}

\textbf{Experiment 5} 
We evaluate 12 sequence design methods introduced in Section~\ref{subsec:methods} on both DNA (UTR) and protein (GFP) sequence design tasks. 
Specifically, for UTR, we leverage two reference datasets: (1) the entire UTR task dataset (280,000 sequences), and (2) the more general GENCODE database~\citet{harrow2012gencode}.\footnote{GENCODE identifies and classifies gene features in human and mouse genomes. Since the UTR task is designed for humans, we create a reference dataset by compiling all human chromosomes annotated with UTRs that are of lengths between 40 and 60 (to ensure comparability with the task-specified 50-length UTRs). This results in a final dataset of 34,060 sequences.}
The protein sequences designed for GFP are evaluated using the five protein measures against a reference dataset of the entire GFP task dataset (51,715).\footnote{A more generic reference protein sequence dataset is not appropriate for this setting, since valid GFP sequences must possess specific properties. For instance, GFP has an unusual covalent bond in its chromophoric Tyr residue, making its functional site's physicochemical microenvironment likely to be quite different from other proteins. In fact, classical atomic simulations designed for general proteins need adjustments to accurately estimate the energetics of GFP \cite{breyfogle2023molecular}. Consequently, measures and metrics derived from or applicable to general proteins might not accurately represent the unique chemical environment of GFP.}
Since the task dataset already contains valid GFP sequences, we can directly compare the generated sequences to those in the dataset, ensuring meaningful property ranges.

\textbf{Results 5} In Figure \ref{fig:validity} we classify the valid (solid) and invalid (shaded) generated UTR and GFP sequences respectively for each of the 12 design methods by applying our proposed suite of biological measures to the 128 \textit{de novo} generated batch. The percentage of valid sequences per method is displayed at the top of each bar. 

For UTR sequences, the left plot presents valid sequences with respect to the UTR task dataset, while the centre plot is with respect to the more general GENCODE human UTR dataset. 
For the task dataset, the methods that generate a higher proportion of biologically meaningful sequences are Auto. CbAS, CbAS, BO-qEI, REINFORCE, and MINs (in descending order), with over 90\% of sequences in their generated batch being valid. COMs obtains a batch with 64.7\% valid sequences, and the remaining methods, GA, GA Mean, GA Min, BootGen, CMA-ES, and GFN-AL (in descending performance order), have less than 30\% valid sequences, with strikingly GFN-AL generating 0\% valid sequences.
This highlights that these methods generate sequences with biological properties significantly different from the task reference dataset. Consequently, we can assume that these methods generate out-of-distribution (OOD) sequences compared to the task dataset distribution. Given the oracle's poor generalization to OOD sequences, it is worth considering whether invalid sequences should be included in the final batch when reporting a method's performance. 
With respect to the GENCODE reference dataset, the top-performing methods are Auto. CbAS, CbAS, BO-qEI, COMs, MINs, and REINFORCE (in descending order), similar to the task dataset results. However, these six methods exhibit much lower performance when evaluated against the GENCODE dataset compared to the task dataset. 
It is also interesting to note that BootGen and CMA-ES are more competitive than GA and its variations on this dataset, however, GFN-AL performs consistently poor, achieving 0\% valid sequences.

For the generated protein sequences, 9 out of 12 methods achieve over 85\% valid GFP sequences in the batch, including COMs, CbAS, Auto. CbAS, GA Min, GA Mean, REINFORCE, BootGen, GA, and MINs, in descending order. Additionally, GFN-AL exhibits poor performance at 5.5\% valid sequences, and the remaining methods, Bo-qEI, CMA-ES, GA and its variations, have no valid sequences; this is not surprising as all three methods obtain very low performance when scored by the oracle model. 

An alternative approach to determining the biophysical validity of \textit{de novo} sequences is to compute the distributional conformity score (DCS) of the sequences to a reference dataset \cite{frey2023protein}. Figure \ref{fig:validity_conf} (Appendix) shows the percentage of valid sequences per method for each task and reference dataset computed using the DCS of each sequence. Comparing these results with Figure \ref{fig:validity}, we observe a similar trend between the methods that consistently generate a high percentage of valid sequences, supporting our evaluation approach and results.

\section{Conclusion}
Our work examines the reliability and consistency of the \textit{in silico} biological sequence design benchmarks. Sequence design methods are commonly evaluated using an ML oracle model, trained a limited dataset of sequences, to score \textit{de novo} generated sequences.

Our first contribution demonstrates that oracles with different seeded runs and architectures result in conflicting rankings of 12 sequence design methods. This lack of consensus among oracles raises concerns regarding the reliability of the oracle models, and our analysis suggests their poor generalisation to out-of-distribution sequences as a key limitation.
Our second contribution introduces a set of biophysical measures to supplement the evaluation procedure. These metrics assess the biological feasibility of \textit{de novo} sequences and effectively limit the space of out-of-distribution sequences the oracle needs to score, thereby improving the robustness of the design procedure. 

In summary, our work highlights the potential limitations in the current evaluation procedure and presents biologically grounded measures to improve the robustness of design benchmarks, with the ultimate goal of enhancing \textit{in silico} design methods. The most significant and challenging direction for future work lies in improving the oracles. With the emergence of more accurate nucleotide \cite{dalla2023nucleotide} and protein language models \cite{lin2022language}, they should be considered for fine-tuning and application as task-specific oracles.
Additionally, the introduced biophysical measures are generic and thus, applicable to all DNA or protein tasks. While these help filter out implausible sequences, developing more accurate and task-specific measures can further refine the state space and increase the likelihood of generating successful, biologically fit sequences. We leave this exploration for future work.


\bibliographystyle{icml2024}

\newpage
\appendix
\onecolumn
\section{Score Distributions of Sequence Datasets}
\begin{figure*}[h!]
    \centering
    \includegraphics[width=0.32\linewidth]{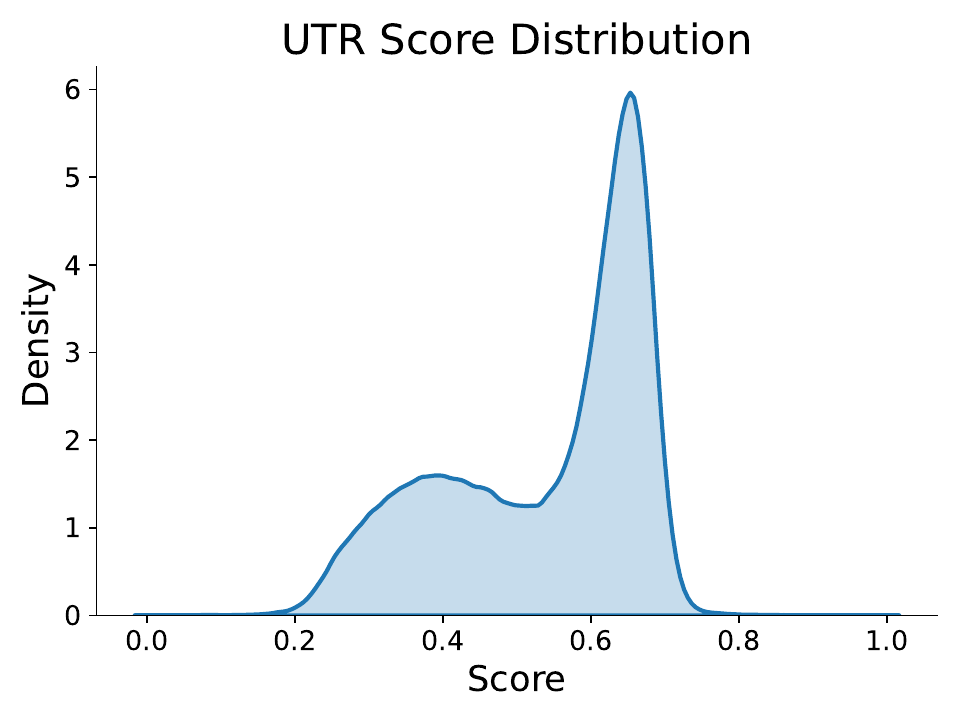}
    \includegraphics[width=0.32\linewidth]{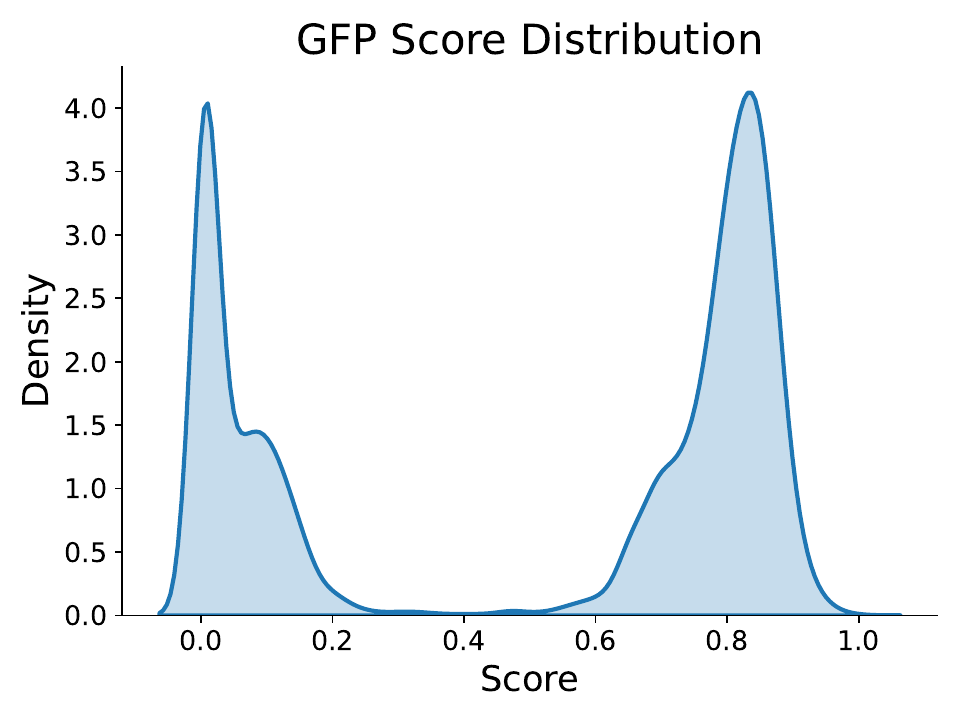}
    \includegraphics[width=0.32\linewidth]{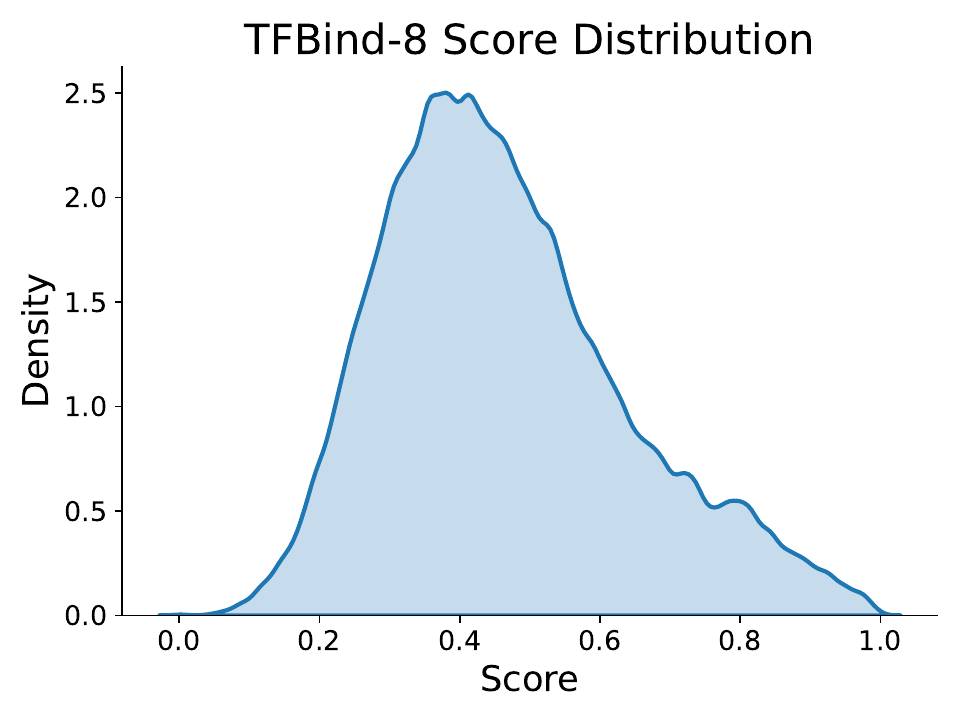}
    \caption{
    The distribution of scores for the UTR (left), GFP (centre), and TFBind-8 (right) datasets.}
    \label{fig:score_distributions}
\end{figure*}

Figure \ref{fig:score_distributions} illustrates the distribution of the scores corresponding to each sequence in the following datasets: UTR (left), GFP (centre), and TFBind-8 (right). 

\section{Additional Sequence Design Results under the Design Bench Oracle}
To assess the robustness of \textit{in silico} design methods, we trained 12 design methods (described in Section \ref{subsec:methods}) using 8 different seeds for both the UTR and GFP tasks, and evaluate them using the Design Bench oracle model. The results, presented in Table \ref{tab:methods_robustness}, show the rankings of the 12 methods for (a) UTR and (b) GFP tasks with 8 seeded replications. The inconsistency in the rankings suggests that these methods are sensitive to the choice of training seed.

\input{tabs/app_exp1}

\begin{wrapfigure}{r!}{0.5\textwidth}
    \vspace{-1\baselineskip}
  \begin{center}
    \includegraphics[width=0.99\linewidth]{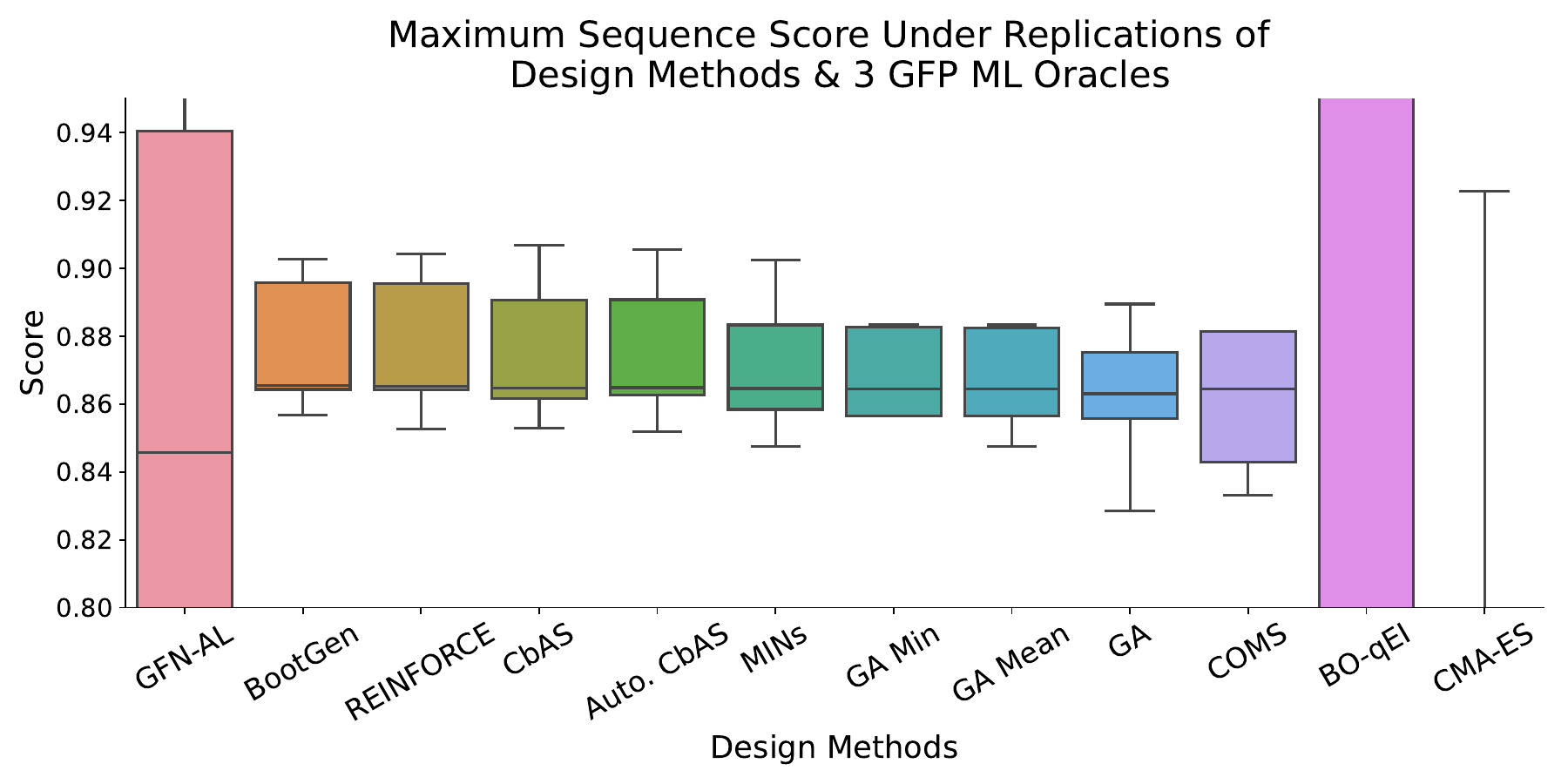}
  \end{center}
   \vspace{-1\baselineskip}
  \caption{Distribution of the maximum score in the batch of \textit{de novo} sequences generated under 8 replications of the design methods and 3 different GFP oracles.}
    \label{fig:gfp_method_score_distribution}
\end{wrapfigure}

\begin{figure*}[h!]
    \vspace{1\baselineskip}
    \centering
    \includegraphics[width=0.99\linewidth]{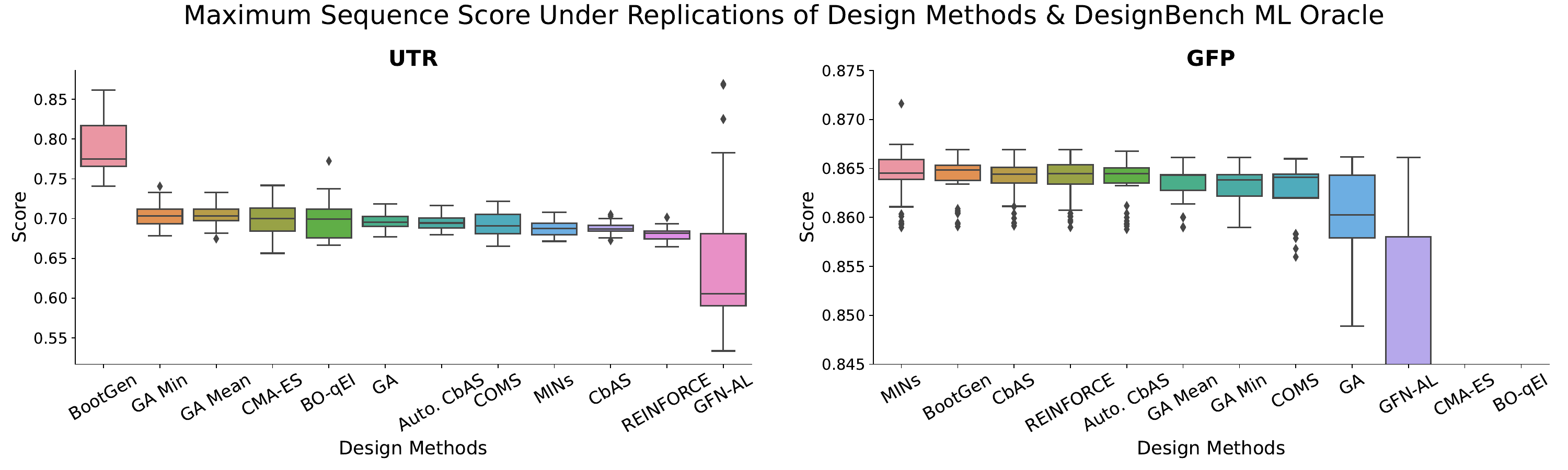}
    \caption{Distribution of the maximum score in the batch of \textit{de novo} sequences generated under 8 replications of the design methods and 5 replications of the DesignBench ML oracles.}
    \label{fig:method_score_distribution}
\end{figure*}

Additionally, to understand how random seeded replications of both the design methods and the oracles impact the maximum score in the final batch of \textit{de novo} generated sequences, we plot the distribution of the maximum score under these replications. Specifically, Figure \ref{fig:gfp_method_score_distribution} shows the distribution of the maximum score under 8 replications of the design methods and three different GFP oracles: DesignBench \cite{designbench_trabucco}, ESM-1b \cite{esm1b_rives}, and TAPE \cite{pex}. Figure \ref{fig:method_score_distribution} depicts the distribution of the maximum score under 8 replications of the design methods and 5 replications of the DesignBench ML oracle for both UTR and GFP.
 
\section{\mbox{Additional Results under the Biophysical Measures}}

\begin{figure*}[h!]

    \centering
    \includegraphics[width=0.33\linewidth]{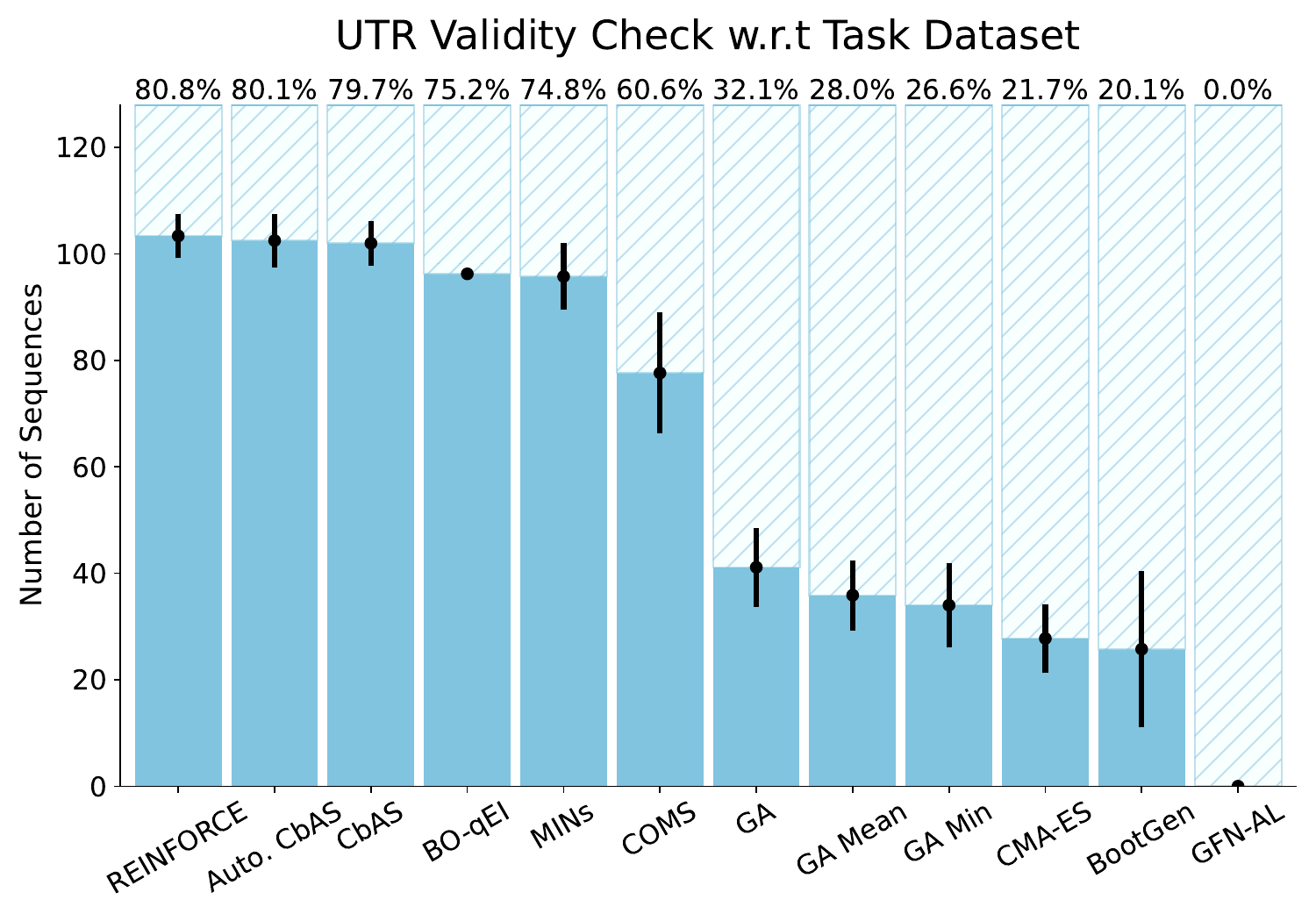}
    \includegraphics[width=0.33\linewidth]{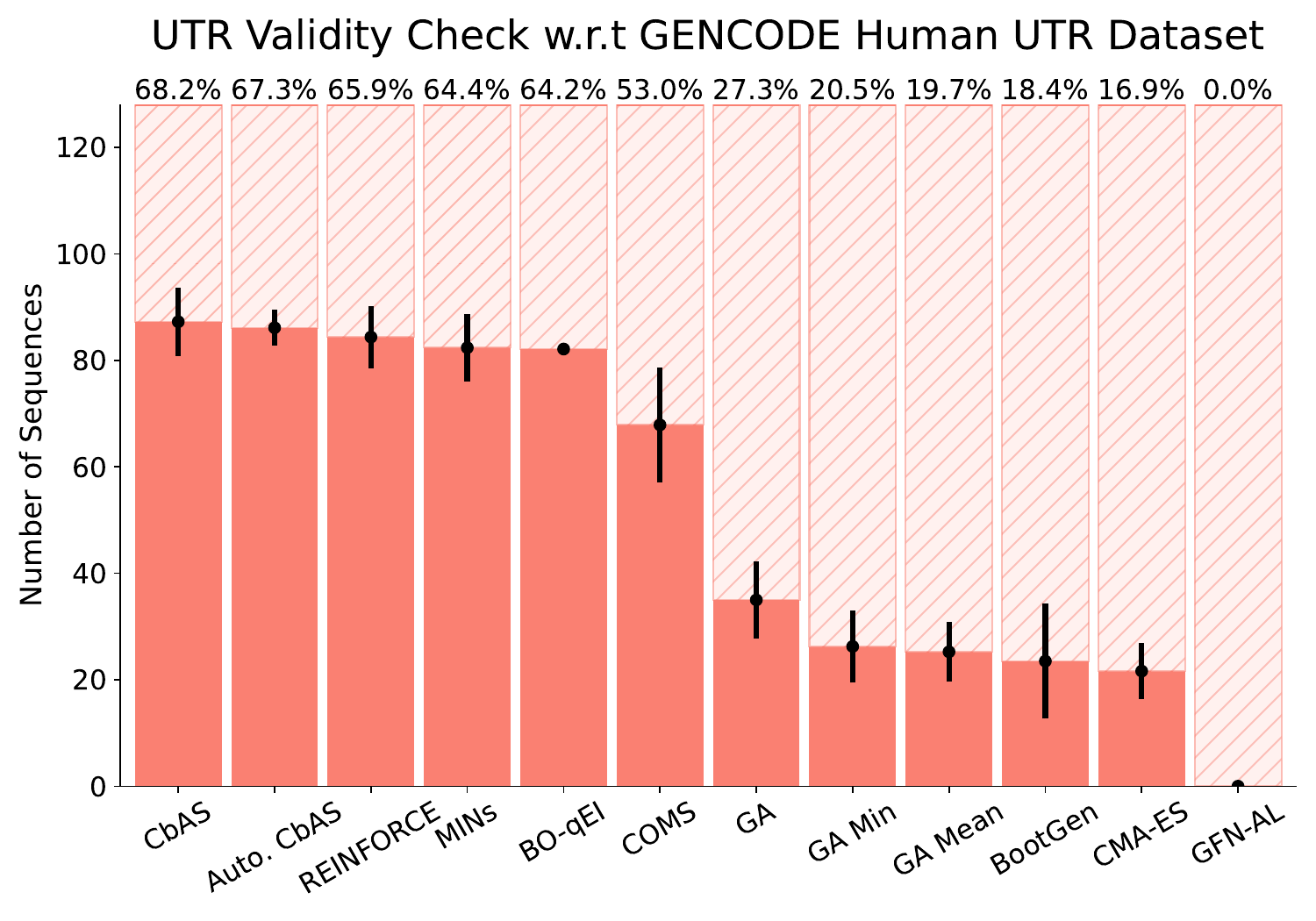}
    \includegraphics[width=0.33\linewidth]{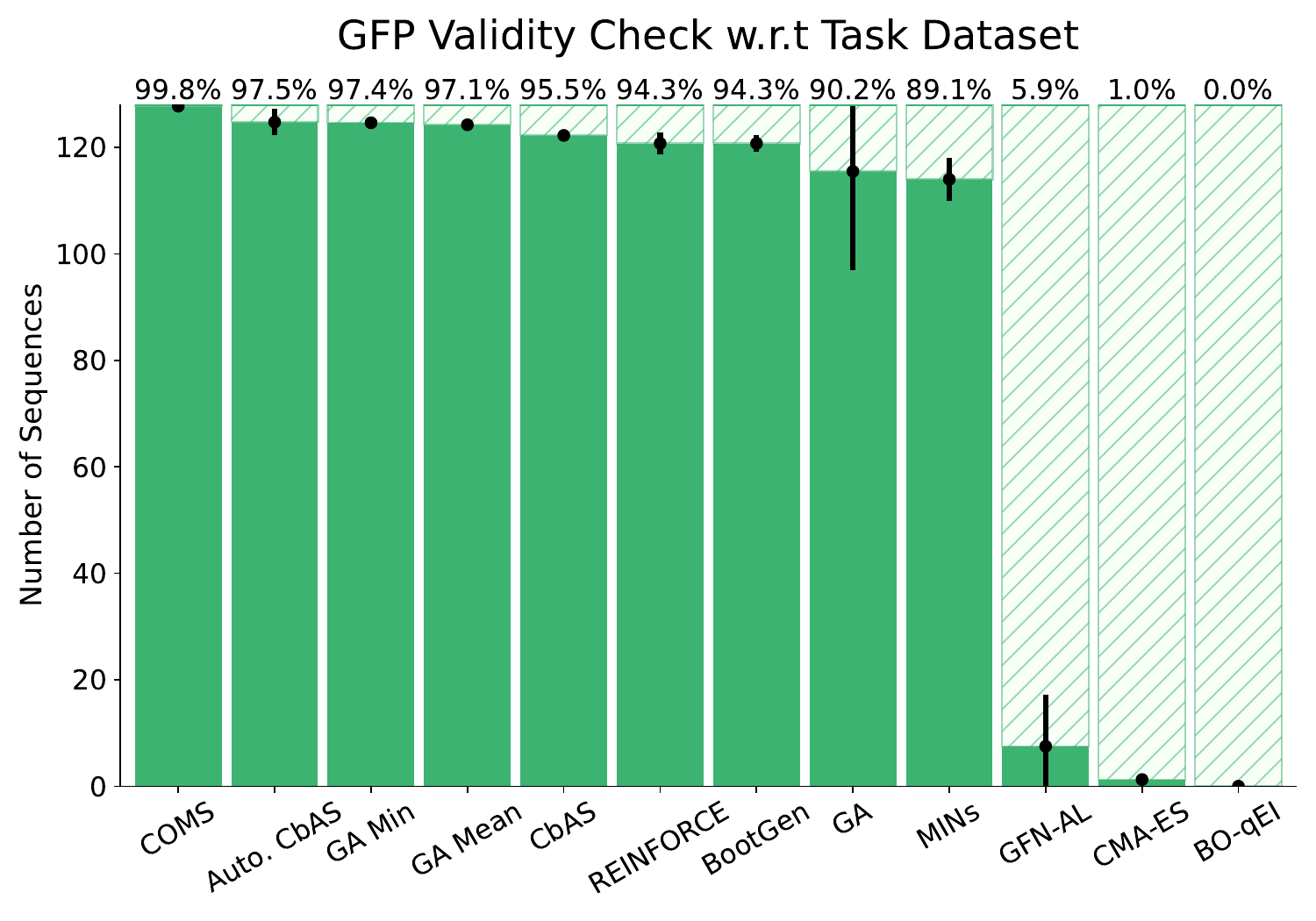}
    \caption{
    Generated sequences classified by DNA (left and centre) and protein (right) suite of biological measures, for 12 sequence design methods with respect to the task dataset (left and right) and the GENCODE database (centre).}
    \label{fig:validity_conf}
\end{figure*}

\citet{frey2023protein} introduced the distributional conformity score (DCS) to improve the quality of \textit{de novo} generated sequences with the aim that the score directly relates to the probability of generating real, biophysically valid proteins. To verify our approach of denoting \textit{de novo} generated sequences as valid, we recompute the validity of the sequences generated by each of the 12 design methods under the DCS. The results, illustrated in Figure \ref{fig:validity_conf}, directly match our results in Figure \ref{fig:validity} for all tasks and reference datasets. 

\end{document}

%% file: tabs/experiment1.tex
\begin{table*}[t!]
  \caption{Relative ranking of 12 sequence design methods (descending order) across five random seed replications of the ML-oracle.}
  \vspace{1ex}
  \label{tab:gfp_seeds}
    \begin{minipage}{.48\textwidth}
      \centering
      \begin{adjustbox}{width=\linewidth,center}
      \begin{tabular}{|c|c|c|c|c|}
        \toprule
            \textbf{Seed 1}                                           & \textbf{Seed 2}                                        & \textbf{Seed 3}                                          & \textbf{Seed 4}                    & \textbf{Seed 5}                                        \\ \midrule 
            \cellcolor[HTML]{E6B8AF}BootGen & \cellcolor[HTML]{E6B8AF}BootGen & \cellcolor[HTML]{E6B8AF}BootGen & \cellcolor[HTML]{E6B8AF}BootGen & \cellcolor[HTML]{E6B8AF}BootGen \\ 
\cellcolor[HTML]{EAD1DC}CMA-ES & \cellcolor[HTML]{CFE2F3}GA Min & \cellcolor[HTML]{CFE2F3}GA Min & \cellcolor[HTML]{CFE2F3}GA Min & \cellcolor[HTML]{F4CCCC}GA Mean  \\ 
\cellcolor[HTML]{F4CCCC}GA Mean  & \cellcolor[HTML]{B5DDCA}BO-qEI & \cellcolor[HTML]{F4CCCC}GA Mean  & \cellcolor[HTML]{F4CCCC}GA Mean  & \cellcolor[HTML]{EAD1DC}CMA-ES \\ 
\cellcolor[HTML]{CFE2F3}GA Min & \cellcolor[HTML]{F4CCCC}GA Mean  & \cellcolor[HTML]{D9D2E9}GA & \cellcolor[HTML]{D9D2E9}GA & \cellcolor[HTML]{B5DDCA}BO-qEI \\ 
\cellcolor[HTML]{C9DAF8}COMs  & \cellcolor[HTML]{D9D2E9}GA & \cellcolor[HTML]{D9EAD3}Auto. CbAS & \cellcolor[HTML]{B5DDCA}BO-qEI & \cellcolor[HTML]{CFE2F3}GA Min \\ 
\cellcolor[HTML]{D9EAD3}Auto. CbAS & \cellcolor[HTML]{D9EAD3}Auto. CbAS & \cellcolor[HTML]{EAD1DC}CMA-ES & \cellcolor[HTML]{EAD1DC}CMA-ES & \cellcolor[HTML]{D9EAD3}Auto. CbAS \\ 
\cellcolor[HTML]{D9D2E9}GA & \cellcolor[HTML]{C9DAF8}COMs  & \cellcolor[HTML]{B5DDCA}BO-qEI & \cellcolor[HTML]{D9EAD3}Auto. CbAS & \cellcolor[HTML]{C9DAF8}COMs  \\ 
\cellcolor[HTML]{B5DDCA}BO-qEI & \cellcolor[HTML]{EAD1DC}CMA-ES & \cellcolor[HTML]{C9DAF8}COMs  & \cellcolor[HTML]{C9DAF8}COMs  & \cellcolor[HTML]{D9D2E9}GA \\ 
\cellcolor[HTML]{FCE5CD}CbAS & \cellcolor[HTML]{D0E0E3}MINs & \cellcolor[HTML]{D0E0E3}MINs & \cellcolor[HTML]{D0E0E3}MINs & \cellcolor[HTML]{FCE5CD}CbAS \\ 
\cellcolor[HTML]{D0E0E3}MINs & \cellcolor[HTML]{FCE5CD}CbAS & \cellcolor[HTML]{FCE5CD}CbAS & \cellcolor[HTML]{FCE5CD}CbAS & \cellcolor[HTML]{D0E0E3}MINs \\ 
\cellcolor[HTML]{FFF2CC}REINFORCE & \cellcolor[HTML]{FFF2CC}REINFORCE & \cellcolor[HTML]{FFF2CC}REINFORCE & \cellcolor[HTML]{FFF2CC}REINFORCE & \cellcolor[HTML]{FFF2CC}REINFORCE \\ 
\cellcolor[HTML]{F8E3A6}GFN-AL  & \cellcolor[HTML]{F8E3A6}GFN-AL  & \cellcolor[HTML]{F8E3A6}GFN-AL  & \cellcolor[HTML]{F8E3A6}GFN-AL  & \cellcolor[HTML]{F8E3A6}GFN-AL  \\

            \bottomrule
            \end{tabular}
            \end{adjustbox}
      \caption*{(a) UTR Design Bench oracle}
      \label{tab:utr_rankings}
    \end{minipage}%
    \hfill
    \begin{minipage}{.48\textwidth}
      \centering
      \begin{adjustbox}{width=\linewidth,center}
      \begin{tabular}{|c|c|c|c|c|}
        \toprule 
            \textbf{Seed 1}                                                 & \textbf{Seed 2}                                          & \textbf{Seed 3}                                          & \textbf{Seed 4}                            & \textbf{Seed 5}                         \\ \midrule                      
            \cellcolor[HTML]{E6B8AF}BootGen & \cellcolor[HTML]{FCE5CD}CbAS & \cellcolor[HTML]{D0E0E3}MINs & \cellcolor[HTML]{D0E0E3}MINs & \cellcolor[HTML]{D0E0E3}MINs \\ 
\cellcolor[HTML]{FFF2CC}REINFORCE & \cellcolor[HTML]{E6B8AF}BootGen & \cellcolor[HTML]{E6B8AF}BootGen & \cellcolor[HTML]{E6B8AF}BootGen & \cellcolor[HTML]{E6B8AF}BootGen \\ 
\cellcolor[HTML]{D0E0E3}MINs & \cellcolor[HTML]{FFF2CC}REINFORCE & \cellcolor[HTML]{FCE5CD}CbAS & \cellcolor[HTML]{D9EAD3}Auto. CbAS & \cellcolor[HTML]{D9EAD3}Auto. CbAS \\ 
\cellcolor[HTML]{D9EAD3}Auto. CbAS & \cellcolor[HTML]{CFE2F3}GA Min & \cellcolor[HTML]{D9EAD3}Auto. CbAS & \cellcolor[HTML]{FFF2CC}REINFORCE & \cellcolor[HTML]{FFF2CC}REINFORCE \\ 
\cellcolor[HTML]{FCE5CD}CbAS & \cellcolor[HTML]{D0E0E3}MINs & \cellcolor[HTML]{FFF2CC}REINFORCE & \cellcolor[HTML]{FCE5CD}CbAS & \cellcolor[HTML]{FCE5CD}CbAS \\ 
\cellcolor[HTML]{C9DAF8}COMs  & \cellcolor[HTML]{F4CCCC}GA Mean  & \cellcolor[HTML]{F4CCCC}GA Mean  & \cellcolor[HTML]{F4CCCC}GA Mean  & \cellcolor[HTML]{F4CCCC}GA Mean  \\ 
\cellcolor[HTML]{F4CCCC}GA Mean  & \cellcolor[HTML]{D9EAD3}Auto. CbAS & \cellcolor[HTML]{CFE2F3}GA Min & \cellcolor[HTML]{CFE2F3}GA Min & \cellcolor[HTML]{CFE2F3}GA Min \\ 
\cellcolor[HTML]{CFE2F3}GA Min & \cellcolor[HTML]{C9DAF8}COMs  & \cellcolor[HTML]{C9DAF8}COMs  & \cellcolor[HTML]{C9DAF8}COMs  & \cellcolor[HTML]{C9DAF8}COMs  \\ 
\cellcolor[HTML]{D9D2E9}GA & \cellcolor[HTML]{D9D2E9}GA & \cellcolor[HTML]{D9D2E9}GA & \cellcolor[HTML]{D9D2E9}GA & \cellcolor[HTML]{D9D2E9}GA \\ 
\cellcolor[HTML]{F8E3A6}GFN-AL  & \cellcolor[HTML]{F8E3A6}GFN-AL  & \cellcolor[HTML]{F8E3A6}GFN-AL  & \cellcolor[HTML]{F8E3A6}GFN-AL  & \cellcolor[HTML]{F8E3A6}GFN-AL  \\ 
\cellcolor[HTML]{EAD1DC}CMA-ES & \cellcolor[HTML]{EAD1DC}CMA-ES & \cellcolor[HTML]{B5DDCA}BO-qEI & \cellcolor[HTML]{EAD1DC}CMA-ES & \cellcolor[HTML]{EAD1DC}CMA-ES \\ 
\cellcolor[HTML]{B5DDCA}BO-qEI & \cellcolor[HTML]{B5DDCA}BO-qEI & \cellcolor[HTML]{EAD1DC}CMA-ES & \cellcolor[HTML]{B5DDCA}BO-qEI & \cellcolor[HTML]{B5DDCA}BO-qEI \\ 
 \bottomrule  
            \end{tabular}
            \end{adjustbox}
      \caption*{(b) GFP Design Bench oracle}
      \label{tab:gfp_rankings}
    \end{minipage}
\end{table*}

%% file: tabs/gfp_experiment2.tex
\begin{table}[t!]
    \centering
    \caption{The ranking of 12 design methods (descending order) for three different GFP oracles: Design Bench Transformer, TAPE, and fine-tuned ESM1b.}
    \vspace{1ex}
    \label{tab:gfp_oracles}
    \begin{adjustbox}{width=0.6\linewidth,center}
    \begin{tabular}{|c|c|c|}
\toprule
\textbf{Design Bench}                                           & \textbf{TAPE}                                          & \textbf{ESM1b}                                                \\ \midrule
\cellcolor[HTML]{E6B8AF}BootGen & \cellcolor[HTML]{E6B8AF}BootGen & \cellcolor[HTML]{F8E3A6}GFN-AL  \\ 
\cellcolor[HTML]{FFF2CC}REINFORCE & \cellcolor[HTML]{FFF2CC}REINFORCE & \cellcolor[HTML]{B5DDCA}BO-qEI \\ 
\cellcolor[HTML]{D0E0E3}MINs & \cellcolor[HTML]{FCE5CD}CbAS & \cellcolor[HTML]{E6B8AF}BootGen \\ 
\cellcolor[HTML]{D9EAD3}Auto. CbAS & \cellcolor[HTML]{D9EAD3}Auto. CbAS & \cellcolor[HTML]{FFF2CC}REINFORCE \\ 
\cellcolor[HTML]{FCE5CD}CbAS & \cellcolor[HTML]{D0E0E3}MINs & \cellcolor[HTML]{FCE5CD}CbAS \\ 
\cellcolor[HTML]{C9DAF8}COMs  & \cellcolor[HTML]{CFE2F3}GA Min & \cellcolor[HTML]{D9EAD3}Auto. CbAS \\ 
\cellcolor[HTML]{F4CCCC}GA Mean  & \cellcolor[HTML]{F4CCCC}GA Mean  & \cellcolor[HTML]{CFE2F3}GA Min \\ 
\cellcolor[HTML]{CFE2F3}GA Min & \cellcolor[HTML]{C9DAF8}COMs  & \cellcolor[HTML]{D0E0E3}MINs \\ 
\cellcolor[HTML]{D9D2E9}GA & \cellcolor[HTML]{D9D2E9}GA & \cellcolor[HTML]{F4CCCC}GA Mean  \\ 
\cellcolor[HTML]{F8E3A6}GFN-AL  & \cellcolor[HTML]{F8E3A6}GFN-AL  & \cellcolor[HTML]{D9D2E9}GA \\ 
\cellcolor[HTML]{EAD1DC}CMA-ES & \cellcolor[HTML]{EAD1DC}CMA-ES & \cellcolor[HTML]{C9DAF8}COMs  \\ 
\cellcolor[HTML]{B5DDCA}BO-qEI & \cellcolor[HTML]{B5DDCA}BO-qEI & \cellcolor[HTML]{EAD1DC}CMA-ES \\ 

  \bottomrule
\end{tabular}
\end{adjustbox}
\vspace{-2ex}
\end{table}

%% file: tabs/app_exp1.tex
\begin{table*}[h!]
  \caption{Relative ranking of 12 sequence design methods (descending order) across 8 random seed replications of the methods, evaluated under the Design Bench oracle.}
  \vspace{1ex}
  \label{tab:methods_robustness}
  \centering
    \begin{minipage}{.9\textwidth}
      \centering
      \caption*{(a) UTR }
      \begin{adjustbox}{width=\linewidth,center}
      \begin{tabular}{|c|c|c|c|c|c|c|c|}
        \toprule
            \textbf{Seed 1}                                           & \textbf{Seed 2}                                        & \textbf{Seed 3}                                          & \textbf{Seed 4}                    & \textbf{Seed 5} & \textbf{Seed 6} & \textbf{Seed 7} & \textbf{Seed 8}                                      \\ \midrule 
            \cellcolor[HTML]{F8E3A6}GFN-AL  & \cellcolor[HTML]{E6B8AF}BootGen & \cellcolor[HTML]{F8E3A6}GFN-AL  & \cellcolor[HTML]{E6B8AF}BootGen & \cellcolor[HTML]{E6B8AF}BootGen & \cellcolor[HTML]{E6B8AF}BootGen & \cellcolor[HTML]{E6B8AF}BootGen & \cellcolor[HTML]{E6B8AF}BootGen \\ 
\cellcolor[HTML]{E6B8AF}BootGen & \cellcolor[HTML]{EAD1DC}CMA-ES & \cellcolor[HTML]{E6B8AF}BootGen & \cellcolor[HTML]{CFE2F3}GA Min & \cellcolor[HTML]{F4CCCC}GA Mean  & \cellcolor[HTML]{EAD1DC}CMA-ES & \cellcolor[HTML]{CFE2F3}GA Min & \cellcolor[HTML]{C9DAF8}COMs  \\ 
\cellcolor[HTML]{EAD1DC}CMA-ES & \cellcolor[HTML]{B5DDCA}BO-QEI & \cellcolor[HTML]{EAD1DC}CMA-ES & \cellcolor[HTML]{B5DDCA}BO-QEI & \cellcolor[HTML]{CFE2F3}GA Min & \cellcolor[HTML]{F4CCCC}GA Mean  & \cellcolor[HTML]{EAD1DC}CMA-ES & \cellcolor[HTML]{CFE2F3}GA Min \\ 
\cellcolor[HTML]{C9DAF8}COMs  & \cellcolor[HTML]{D9EAD3}Auto. CbAS & \cellcolor[HTML]{D9EAD3}Auto. CbAS & \cellcolor[HTML]{FCE5CD}CbAS & \cellcolor[HTML]{B5DDCA}BO-QEI & \cellcolor[HTML]{C9DAF8}COMs  & \cellcolor[HTML]{F4CCCC}GA Mean  & \cellcolor[HTML]{F4CCCC}GA Mean  \\ 
\cellcolor[HTML]{F4CCCC}GA Mean  & \cellcolor[HTML]{C9DAF8}COMs  & \cellcolor[HTML]{F4CCCC}GA Mean  & \cellcolor[HTML]{D9EAD3}Auto. CbAS & \cellcolor[HTML]{EAD1DC}CMA-ES & \cellcolor[HTML]{B5DDCA}BO-QEI & \cellcolor[HTML]{FFF2CC}REINFORCE & \cellcolor[HTML]{EAD1DC}CMA-ES \\ 
\cellcolor[HTML]{D9D2E9}GA & \cellcolor[HTML]{CFE2F3}GA Min & \cellcolor[HTML]{FCE5CD}CbAS & \cellcolor[HTML]{D0E0E3}MINs & \cellcolor[HTML]{D0E0E3}MINs & \cellcolor[HTML]{D9D2E9}GA & \cellcolor[HTML]{D0E0E3}MINs & \cellcolor[HTML]{D9D2E9}GA \\ 
\cellcolor[HTML]{D0E0E3}MINs & \cellcolor[HTML]{D9D2E9}GA & \cellcolor[HTML]{D9D2E9}GA & \cellcolor[HTML]{C9DAF8}COMs  & \cellcolor[HTML]{D9D2E9}GA & \cellcolor[HTML]{CFE2F3}GA Min & \cellcolor[HTML]{FCE5CD}CbAS & \cellcolor[HTML]{FFF2CC}REINFORCE \\ 
\cellcolor[HTML]{CFE2F3}GA Min & \cellcolor[HTML]{FCE5CD}CbAS & \cellcolor[HTML]{B5DDCA}BO-QEI & \cellcolor[HTML]{F4CCCC}GA Mean  & \cellcolor[HTML]{D9EAD3}Auto. CbAS & \cellcolor[HTML]{D9EAD3}Auto. CbAS & \cellcolor[HTML]{D9EAD3}Auto. CbAS & \cellcolor[HTML]{D9EAD3}Auto. CbAS \\ 
\cellcolor[HTML]{FCE5CD}CbAS & \cellcolor[HTML]{D0E0E3}MINs & \cellcolor[HTML]{FFF2CC}REINFORCE & \cellcolor[HTML]{D9D2E9}GA & \cellcolor[HTML]{FCE5CD}CbAS & \cellcolor[HTML]{FCE5CD}CbAS & \cellcolor[HTML]{D9D2E9}GA & \cellcolor[HTML]{D0E0E3}MINs \\ 
\cellcolor[HTML]{D9EAD3}Auto. CbAS & \cellcolor[HTML]{F4CCCC}GA Mean  & \cellcolor[HTML]{CFE2F3}GA Min & \cellcolor[HTML]{EAD1DC}CMA-ES & \cellcolor[HTML]{C9DAF8}COMs  & \cellcolor[HTML]{FFF2CC}REINFORCE & \cellcolor[HTML]{C9DAF8}COMs  & \cellcolor[HTML]{FCE5CD}CbAS \\ 
\cellcolor[HTML]{B5DDCA}BO-QEI & \cellcolor[HTML]{FFF2CC}REINFORCE & \cellcolor[HTML]{D0E0E3}MINs & \cellcolor[HTML]{FFF2CC}REINFORCE & \cellcolor[HTML]{FFF2CC}REINFORCE & \cellcolor[HTML]{D0E0E3}MINs & \cellcolor[HTML]{B5DDCA}BO-QEI & \cellcolor[HTML]{B5DDCA}BO-QEI \\ 
\cellcolor[HTML]{FFF2CC}REINFORCE & \cellcolor[HTML]{F8E3A6}GFN-AL  & \cellcolor[HTML]{C9DAF8}COMs  & \cellcolor[HTML]{F8E3A6}GFN-AL  & \cellcolor[HTML]{F8E3A6}GFN-AL  & \cellcolor[HTML]{F8E3A6}GFN-AL  & \cellcolor[HTML]{F8E3A6}GFN-AL  & \cellcolor[HTML]{F8E3A6}GFN-AL  \\
            \bottomrule
            \end{tabular}
            \end{adjustbox}
      
      \label{tab:utr_methos}
    \end{minipage}%
    \hfill
    \centering
    \begin{minipage}{.9\textwidth}
      \centering
      \caption*{(b) GFP}
      \begin{adjustbox}{width=\linewidth,center}
      \begin{tabular}{|c|c|c|c|c|c|c|c|}
        \toprule
            \textbf{Seed 1}                                           & \textbf{Seed 2}                                        & \textbf{Seed 3}                                          & \textbf{Seed 4}                    & \textbf{Seed 5} & \textbf{Seed 6} & \textbf{Seed 7} & \textbf{Seed 8}                                      \\ \midrule 
            \cellcolor[HTML]{FCE5CD}CbAS & \cellcolor[HTML]{FFF2CC}REINFORCE & \cellcolor[HTML]{D0E0E3}MINs & \cellcolor[HTML]{E6B8AF}BootGen & \cellcolor[HTML]{FFF2CC}REINFORCE & \cellcolor[HTML]{D9EAD3}Auto. CbAS & \cellcolor[HTML]{D9EAD3}Auto. CbAS & \cellcolor[HTML]{D0E0E3}MINs \\ 
\cellcolor[HTML]{D9EAD3}Auto. CbAS & \cellcolor[HTML]{E6B8AF}BootGen & \cellcolor[HTML]{E6B8AF}BootGen & \cellcolor[HTML]{FCE5CD}CbAS & \cellcolor[HTML]{E6B8AF}BootGen & \cellcolor[HTML]{E6B8AF}BootGen & \cellcolor[HTML]{FFF2CC}REINFORCE & \cellcolor[HTML]{FFF2CC}REINFORCE \\ 
\cellcolor[HTML]{FFF2CC}REINFORCE & \cellcolor[HTML]{FCE5CD}CbAS & \cellcolor[HTML]{FFF2CC}REINFORCE & \cellcolor[HTML]{FFF2CC}REINFORCE & \cellcolor[HTML]{FCE5CD}CbAS & \cellcolor[HTML]{D0E0E3}MINs & \cellcolor[HTML]{D0E0E3}MINs & \cellcolor[HTML]{FCE5CD}CbAS \\ 
\cellcolor[HTML]{E6B8AF}BootGen & \cellcolor[HTML]{D9EAD3}Auto. CbAS & \cellcolor[HTML]{C9DAF8}COMs  & \cellcolor[HTML]{D0E0E3}MINs & \cellcolor[HTML]{D9EAD3}Auto. CbAS & \cellcolor[HTML]{FFF2CC}REINFORCE & \cellcolor[HTML]{C9DAF8}COMs  & \cellcolor[HTML]{E6B8AF}BootGen \\ 
\cellcolor[HTML]{D0E0E3}MINs & \cellcolor[HTML]{D0E0E3}MINs & \cellcolor[HTML]{FCE5CD}CbAS & \cellcolor[HTML]{C9DAF8}COMs  & \cellcolor[HTML]{D0E0E3}MINs & \cellcolor[HTML]{C9DAF8}COMs  & \cellcolor[HTML]{FCE5CD}CbAS & \cellcolor[HTML]{D9EAD3}Auto. CbAS \\ 
\cellcolor[HTML]{C9DAF8}COMs  & \cellcolor[HTML]{C9DAF8}COMs  & \cellcolor[HTML]{D9EAD3}Auto. CbAS & \cellcolor[HTML]{D9EAD3}Auto. CbAS & \cellcolor[HTML]{C9DAF8}COMs  & \cellcolor[HTML]{FCE5CD}CbAS & \cellcolor[HTML]{E6B8AF}BootGen & \cellcolor[HTML]{C9DAF8}COMs  \\ 
\cellcolor[HTML]{F8E3A6}GFN-AL  & \cellcolor[HTML]{F8E3A6}GFN-AL  & \cellcolor[HTML]{F8E3A6}GFN-AL  & \cellcolor[HTML]{EAD1DC}CMA-ES & \cellcolor[HTML]{EAD1DC}CMA-ES & \cellcolor[HTML]{F8E3A6}GFN-AL  & \cellcolor[HTML]{F8E3A6}GFN-AL  & \cellcolor[HTML]{F8E3A6}GFN-AL  \\ 
\cellcolor[HTML]{EAD1DC}CMA-ES & \cellcolor[HTML]{EAD1DC}CMA-ES & \cellcolor[HTML]{EAD1DC}CMA-ES & \cellcolor[HTML]{CFE2F3}GA Min & \cellcolor[HTML]{F4CCCC}GA Mean  & \cellcolor[HTML]{EAD1DC}CMA-ES & \cellcolor[HTML]{EAD1DC}CMA-ES & \cellcolor[HTML]{EAD1DC}CMA-ES \\ 
\cellcolor[HTML]{F4CCCC}GA Mean  & \cellcolor[HTML]{CFE2F3}GA Min & \cellcolor[HTML]{D9D2E9}GA & \cellcolor[HTML]{F4CCCC}GA Mean  & \cellcolor[HTML]{CFE2F3}GA Min & \cellcolor[HTML]{F4CCCC}GA Mean  & \cellcolor[HTML]{D9D2E9}GA & \cellcolor[HTML]{F4CCCC}GA Mean  \\ 
\cellcolor[HTML]{CFE2F3}GA Min & \cellcolor[HTML]{F4CCCC}GA Mean  & \cellcolor[HTML]{F4CCCC}GA Mean  & \cellcolor[HTML]{B5DDCA}BO-qEI & \cellcolor[HTML]{D9D2E9}GA & \cellcolor[HTML]{B5DDCA}BO-qEI & \cellcolor[HTML]{CFE2F3}GA Min & \cellcolor[HTML]{CFE2F3}GA Min \\ 
\cellcolor[HTML]{B5DDCA}BO-qEI & \cellcolor[HTML]{B5DDCA}BO-qEI & \cellcolor[HTML]{CFE2F3}GA Min & \cellcolor[HTML]{F8E3A6}GFN-AL  & \cellcolor[HTML]{F8E3A6}GFN-AL  & \cellcolor[HTML]{CFE2F3}GA Min & \cellcolor[HTML]{B5DDCA}BO-qEI & \cellcolor[HTML]{B5DDCA}BO-qEI \\ 
\cellcolor[HTML]{D9D2E9}GA & \cellcolor[HTML]{D9D2E9}GA & \cellcolor[HTML]{B5DDCA}BO-qEI & \cellcolor[HTML]{D9D2E9}GA & \cellcolor[HTML]{B5DDCA}BO-qEI & \cellcolor[HTML]{D9D2E9}GA & \cellcolor[HTML]{F4CCCC}GA Mean  & \cellcolor[HTML]{D9D2E9}GA \\
            \bottomrule
            \end{tabular}
            \end{adjustbox}
      \label{tab:gfp_methods}
    \end{minipage}
\end{table*}

%% file: example_paper.bbl
\begin{thebibliography}{30}
\providecommand{\natexlab}[1]{#1}
\providecommand{\url}[1]{\texttt{#1}}
\expandafter\ifx\csname urlstyle\endcsname\relax
  \providecommand{\doi}[1]{doi: #1}\else
  \providecommand{\doi}{doi: \begingroup \urlstyle{rm}\Url}\fi

\bibitem[Angermueller et~al.(2019)Angermueller, Dohan, Belanger, Deshpande, Murphy, and Colwell]{angermueller2019model}
Angermueller, C., Dohan, D., Belanger, D., Deshpande, R., Murphy, K., and Colwell, L.
\newblock Model-based reinforcement learning for biological sequence design.
\newblock In \emph{International conference on learning representations}, 2019.

\bibitem[Angermueller et~al.(2020)Angermueller, Belanger, Gane, Mariet, Dohan, Murphy, Colwell, and Sculley]{angermueller2020population}
Angermueller, C., Belanger, D., Gane, A., Mariet, Z., Dohan, D., Murphy, K., Colwell, L., and Sculley, D.
\newblock Population-based black-box optimization for biological sequence design.
\newblock In \emph{International conference on machine learning}, pp.\  324--334. PMLR, 2020.

\bibitem[Breyfogle et~al.(2023)Breyfogle, Blood, Rosnik, and Krueger]{breyfogle2023molecular}
Breyfogle, K.~L., Blood, D.~L., Rosnik, A.~M., and Krueger, B.~P.
\newblock Molecular dynamics force field parameters for the egfp chromophore and some of its analogues.
\newblock \emph{The Journal of Physical Chemistry B}, 127\penalty0 (26):\penalty0 5772--5788, 2023.

\bibitem[Brookes et~al.(2019)Brookes, Park, and Listgarten]{cbas_brookes}
Brookes, D., Park, H., and Listgarten, J.
\newblock Conditioning by adaptive sampling for robust design.
\newblock In \emph{International conference on machine learning}, pp.\  773--782. PMLR, 2019.

\bibitem[Buttenschoen et~al.(2024)Buttenschoen, Morris, and Deane]{posebusters}
Buttenschoen, M., Morris, G.~M., and Deane, C.~M.
\newblock Posebusters: Ai-based docking methods fail to generate physically valid poses or generalise to novel sequences.
\newblock \emph{Chemical Science}, 2024.

\bibitem[Chen et~al.(2023)Chen, Zhang, Liu, and Coates]{chen2023bidirectional}
Chen, C., Zhang, Y., Liu, X., and Coates, M.
\newblock Bidirectional learning for offline model-based biological sequence design.
\newblock In \emph{International Conference on Machine Learning}, pp.\  5351--5366. PMLR, 2023.

\bibitem[Dalla-Torre et~al.(2023)Dalla-Torre, Gonzalez, Mendoza-Revilla, Carranza, Grzywaczewski, Oteri, Dallago, Trop, Sirelkhatim, Richard, et~al.]{dalla2023nucleotide}
Dalla-Torre, H., Gonzalez, L., Mendoza-Revilla, J., Carranza, N.~L., Grzywaczewski, A.~H., Oteri, F., Dallago, C., Trop, E., Sirelkhatim, H., Richard, G., et~al.
\newblock The nucleotide transformer: Building and evaluating robust foundation models for human genomics.
\newblock \emph{bioRxiv}, pp.\  2023--01, 2023.

\bibitem[Fannjiang \& Listgarten(2020)Fannjiang and Listgarten]{autofocused_fannjiang}
Fannjiang, C. and Listgarten, J.
\newblock Autofocused oracles for model-based design.
\newblock \emph{Advances in Neural Information Processing Systems}, 33:\penalty0 12945--12956, 2020.

\bibitem[Frey et~al.(2024)Frey, Berenberg, Zadorozhny, Kleinhenz, Lafrance-Vanasse, Hotzel, Wu, Ra, Bonneau, Cho, et~al.]{frey2023protein}
Frey, N.~C., Berenberg, D., Zadorozhny, K., Kleinhenz, J., Lafrance-Vanasse, J., Hotzel, I., Wu, Y., Ra, S., Bonneau, R., Cho, K., et~al.
\newblock Protein discovery with discrete walk-jump sampling.
\newblock \emph{International Conference on Learning Representations}, 2024.

\bibitem[Hansen(2006)]{hansen2006cma}
Hansen, N.
\newblock The cma evolution strategy: a comparing review.
\newblock \emph{Towards a new evolutionary computation: Advances in the estimation of distribution algorithms}, pp.\  75--102, 2006.

\bibitem[Harris et~al.(2023)Harris, Didi, Jamasb, Joshi, Mathis, Lio, and Blundell]{posecheck}
Harris, C., Didi, K., Jamasb, A.~R., Joshi, C.~K., Mathis, S.~V., Lio, P., and Blundell, T.
\newblock Benchmarking generated poses: How rational is structure-based drug design with generative models?
\newblock \emph{arXiv preprint arXiv:2308.07413}, 2023.

\bibitem[Harrow et~al.(2012)Harrow, Frankish, Gonzalez, Tapanari, Diekhans, Kokocinski, Aken, Barrell, Zadissa, Searle, et~al.]{harrow2012gencode}
Harrow, J., Frankish, A., Gonzalez, J.~M., Tapanari, E., Diekhans, M., Kokocinski, F., Aken, B.~L., Barrell, D., Zadissa, A., Searle, S., et~al.
\newblock Gencode: the reference human genome annotation for the encode project.
\newblock \emph{Genome research}, 22\penalty0 (9):\penalty0 1760--1774, 2012.

\bibitem[Jain et~al.(2022)Jain, Bengio, Hernandez-Garcia, Rector-Brooks, Dossou, Ekbote, Fu, Zhang, Kilgour, Zhang, et~al.]{gflownets}
Jain, M., Bengio, E., Hernandez-Garcia, A., Rector-Brooks, J., Dossou, B.~F., Ekbote, C.~A., Fu, J., Zhang, T., Kilgour, M., Zhang, D., et~al.
\newblock Biological sequence design with gflownets.
\newblock In \emph{International Conference on Machine Learning}, pp.\  9786--9801. PMLR, 2022.

\bibitem[Kim et~al.(2024)Kim, Berto, Ahn, and Park]{bootgen}
Kim, M., Berto, F., Ahn, S., and Park, J.
\newblock Bootstrapped training of score-conditioned generator for offline design of biological sequences.
\newblock \emph{Advances in Neural Information Processing Systems}, 36, 2024.

\bibitem[Kumar \& Levine(2020)Kumar and Levine]{mins_kumar}
Kumar, A. and Levine, S.
\newblock Model inversion networks for model-based optimization.
\newblock \emph{Advances in neural information processing systems}, 33:\penalty0 5126--5137, 2020.

\bibitem[Lin et~al.(2022)Lin, Akin, Rao, Hie, Zhu, Lu, Smetanin, dos Santos~Costa, Fazel-Zarandi, Sercu, Candido, et~al.]{lin2022language}
Lin, Z., Akin, H., Rao, R., Hie, B., Zhu, Z., Lu, W., Smetanin, N., dos Santos~Costa, A., Fazel-Zarandi, M., Sercu, T., Candido, S., et~al.
\newblock Language models of protein sequences at the scale of evolution enable accurate structure prediction.
\newblock \emph{bioRxiv}, 2022.

\bibitem[Notin et~al.(2024)Notin, Kollasch, Ritter, Van~Niekerk, Paul, Spinner, Rollins, Shaw, Orenbuch, Weitzman, et~al.]{notin2024proteingym}
Notin, P., Kollasch, A., Ritter, D., Van~Niekerk, L., Paul, S., Spinner, H., Rollins, N., Shaw, A., Orenbuch, R., Weitzman, R., et~al.
\newblock Proteingym: large-scale benchmarks for protein fitness prediction and design.
\newblock \emph{Advances in Neural Information Processing Systems}, 36, 2024.

\bibitem[Rao et~al.(2019)Rao, Bhattacharya, Thomas, Duan, Chen, Canny, Abbeel, and Song]{tape}
Rao, R., Bhattacharya, N., Thomas, N., Duan, Y., Chen, P., Canny, J., Abbeel, P., and Song, Y.
\newblock Evaluating protein transfer learning with tape.
\newblock \emph{Advances in neural information processing systems}, 32, 2019.

\bibitem[Ren et~al.(2022)Ren, Li, Ding, Zhou, Ma, and Peng]{pex}
Ren, Z., Li, J., Ding, F., Zhou, Y., Ma, J., and Peng, J.
\newblock Proximal exploration for model-guided protein sequence design.
\newblock In \emph{International Conference on Machine Learning}, pp.\  18520--18536. PMLR, 2022.

\bibitem[Rives et~al.(2021)Rives, Meier, Sercu, Goyal, Lin, Liu, Guo, Ott, Zitnick, Ma, et~al.]{esm1b_rives}
Rives, A., Meier, J., Sercu, T., Goyal, S., Lin, Z., Liu, J., Guo, D., Ott, M., Zitnick, C.~L., Ma, J., et~al.
\newblock Biological structure and function emerge from scaling unsupervised learning to 250 million protein sequences.
\newblock \emph{Proceedings of the National Academy of Sciences}, 118\penalty0 (15):\penalty0 e2016239118, 2021.

\bibitem[Sample et~al.(2019)Sample, Wang, Reid, Presnyak, McFadyen, Morris, and Seelig]{utr_pjsample}
Sample, P.~J., Wang, B., Reid, D.~W., Presnyak, V., McFadyen, I.~J., Morris, D.~R., and Seelig, G.
\newblock Human 5' utr design and variant effect prediction from a massively parallel translation assay.
\newblock \emph{Nature biotechnology}, 37\penalty0 (7):\penalty0 803--809, 2019.

\bibitem[Sarkisyan et~al.(2016)Sarkisyan, Bolotin, Meer, Usmanova, Mishin, Sharonov, Ivankov, Bozhanova, Baranov, Soylemez, et~al.]{gfp_sarkisyan}
Sarkisyan, K.~S., Bolotin, D.~A., Meer, M.~V., Usmanova, D.~R., Mishin, A.~S., Sharonov, G.~V., Ivankov, D.~N., Bozhanova, N.~G., Baranov, M.~S., Soylemez, O., et~al.
\newblock Local fitness landscape of the green fluorescent protein.
\newblock \emph{Nature}, 533\penalty0 (7603):\penalty0 397--401, 2016.

\bibitem[Song \& Li(2023)Song and Li]{song2023importance}
Song, Z. and Li, L.
\newblock Importance weighted expectation-maximization for protein sequence design.
\newblock In \emph{International Conference on Machine Learning}, pp.\  32349--32364. PMLR, 2023.

\bibitem[Spinner et~al.(2024)Spinner, Kollasch, and Marks]{spinner2024well}
Spinner, H., Kollasch, A.~W., and Marks, D.~S.
\newblock How well do generative protein models generate?
\newblock In \emph{ICLR 2024 Workshop on Generative and Experimental Perspectives for Biomolecular Design}, 2024.

\bibitem[Tagasovska et~al.(2024)Tagasovska, Park, Kirchmeyer, Frey, Watkins, Ismail, Jamasb, Lee, Bryson, Ra, et~al.]{tagasovska2023antibody}
Tagasovska, N., Park, J.~W., Kirchmeyer, M., Frey, N.~C., Watkins, A.~M., Ismail, A.~A., Jamasb, A.~R., Lee, E., Bryson, T., Ra, S., et~al.
\newblock Antibody domainbed: Out-of-distribution generalization in therapeutic protein design.
\newblock \emph{International Conference on Learning Representations}, 2024.

\bibitem[Trabucco et~al.(2021)Trabucco, Kumar, Geng, and Levine]{coms}
Trabucco, B., Kumar, A., Geng, X., and Levine, S.
\newblock Conservative objective models for effective offline model-based optimization.
\newblock In \emph{International Conference on Machine Learning}, pp.\  10358--10368. PMLR, 2021.

\bibitem[Trabucco et~al.(2022)Trabucco, Geng, Kumar, and Levine]{designbench_trabucco}
Trabucco, B., Geng, X., Kumar, A., and Levine, S.
\newblock Design-bench: Benchmarks for data-driven offline model-based optimization.
\newblock In \emph{International Conference on Machine Learning}, pp.\  21658--21676. PMLR, 2022.

\bibitem[Wang et~al.(2023)Wang, Tang, Huang, Pan, Yang, Yang, Mu, and Yang]{wang2023self}
Wang, Y., Tang, H., Huang, L., Pan, L., Yang, L., Yang, H., Mu, F., and Yang, M.
\newblock Self-play reinforcement learning guides protein engineering.
\newblock \emph{Nature Machine Intelligence}, 5\penalty0 (8):\penalty0 845--860, 2023.

\bibitem[Williams(1992)]{reinforce}
Williams, R.~J.
\newblock Simple statistical gradient-following algorithms for connectionist reinforcement learning.
\newblock \emph{Machine learning}, 8:\penalty0 229--256, 1992.

\bibitem[Wilson et~al.(2018)Wilson, Hutter, and Deisenroth]{wilson2018maximizing}
Wilson, J., Hutter, F., and Deisenroth, M.
\newblock Maximizing acquisition functions for bayesian optimization.
\newblock \emph{Advances in neural information processing systems}, 31, 2018.

\end{thebibliography}
